\documentclass[]{fairmeta}

\usepackage{amsmath}
\usepackage{amssymb}
\usepackage{algorithm}
\usepackage{algorithmic}

\title{Predictable Scaling: Efficient Model Exploration via Scaling-Law Prediction}

\author[1]{Aishwarya Wesanekar}
\author[1]{Sheng-Tao Yang}
\author[1]{Jiawei Zang}
\author[1]{Flavio Sales Truzzi}
\author[1]{Chen Chen}
\author[1]{Nader Beigiparast}
\author[1]{Shengyuan Huang}
\author[1]{Roshanak Zilouchian Moghaddam}
\author[1]{Vitor Cid}
\author[1]{Colin Clement}
\author[1]{Vinayak Thapliyal}
\author[1,*]{Sebastian Ament}
\author[1]{Jihao Andreas Lin}
\author[1]{Vincent Huang}
\author[1]{Kai Han}
\author[1]{Boyi Jiang}
\author[1]{Wenlin Chen}
\author[1]{Chaorong Chen}
\author[1]{Ashwin Kumar}
\author[1,*]{Maximilian Balandat}
\author[1]{David Eriksson}
\author[1]{Yifei Zhang}
\author[1]{Pan Zhong}

\affiliation[1]{Meta Platforms, Inc.}
\contribution[*]{Work done at Meta}

\newcommand{\abstracttext}{%
Evaluating candidate changes (hypotheses) to recommendation models requires full-scale training
runs consuming thousands of GPU-hours, severely limiting the hypothesis space engineers can
feasibly explore. We introduce Predictable Scaling (PS), a systematic framework that exploits
empirically verified regularity conditions to predict full-scale model performance from
small-scale proxies. PS constructs scaled-down variants via composable scaling functions along
distinct resource axes, and projects metrics to the target scale via curve fitting with
calibrated uncertainty estimates. We compare three prediction strategies of increasing
expressiveness: even the simplest (raw proxy) achieves prediction error within the empirical
noise floor of repeated-run variance; the most expressive (Gaussian Process) provides an
indicative $15\%$ relative MAE improvement and calibrated uncertainty. Evaluation on two
ranking models of increasing complexity---a task-specific model and a foundation
model---demonstrates that PS correctly ranks actionable hypotheses across diverse hypothesis
types at $79$--$88\%$ compute reductions.%
}

\abstract{\abstracttext}

\correspondence{Aishwarya Wesanekar at \email{awesanekar@meta.com}}

\begin{document}

\maketitle

\section{Introduction}
\label{sec:intro}

In industrial recommendation systems, model exploration---evaluating candidate changes against established baselines---dominates computational requirements. Each candidate requires a full training run, and the candidate set grows multiplicatively with the design space (architectures $\times$ features $\times$ hyperparameters). While scaling laws have been extensively studied for LLMs~\citep{kaplan2020scaling, hoffmann2022training}, their exploitation for predicting hypothesis-level performance differences in recommendation systems remains largely unexplored. Our key empirical observation is that \emph{relative performance gaps} between hypotheses behave as smooth, low-dimensional functions of scale, even when absolute curves are noisy. We present Predictable Scaling (PS), a framework that leverages this regularity to predict hypothesis performance at a fraction of full training compute. Our contributions:

\begin{itemize}
\item \textbf{Formalization and scaling functions.} We formalize reduced-scale evaluation as prediction over relative gaps with explicit regularity conditions, and introduce six composable scaling functions that reduce compute along distinct resource axes (\cref{sec:framework}).
\item \textbf{Predictor comparison with uncertainty.} Three strategies: simplest achieves error within AA noise floor; GP provides calibrated uncertainty for reliability screening (\cref{sec:results}).
\item \textbf{Large-scale evaluations.} PS validated on two ranking models achieving $79$--$88\%$ compute reduction while correctly ranking actionable hypotheses (\cref{sec:results}).
\end{itemize}

\section{Related Work}
\label{sec:related}

\paragraph{Scaling laws in deep learning.} Kaplan et al.~\citep{kaplan2020scaling} and Hoffmann et al.~\citep{hoffmann2022training} established power-law scaling for language models. Gadre et al.~\citep{gadre2024language} show these laws hold under over-training, relevant to PS's under-trained proxy regime. However, these results derive from homogeneous Transformers; recommendation systems break this assumption with heterogeneous architectures coupling sparse embeddings, dense layers, and sequential components.

\paragraph{Efficient candidate evaluation.} Multi-fidelity methods (Hyperband~\citep{li2018hyperband}, BOHB~\citep{falkner2018bohb}) adaptively terminate unpromising configurations within a fixed architecture. $\mu$P~\citep{yang2022mup} achieves zero-shot hyperparameter transfer. PS is complementary: it predicts impact of \emph{structural} changes (architecture, features, data) that each require distinct training runs and cannot be framed as early-stopping decisions.

\paragraph{Small-to-large prediction.} Tay et al.~\citep{tay2022scalinglaws} show best architectures can change across scales; Choshen et al.~\citep{choshen2025hitchhiker} and Lee et al.~\citep{lee2025bayesianscaling} propose methodology and Bayesian extrapolation for LLMs. All predict absolute LLM losses; PS predicts relative gaps between recommendation model hypotheses.

\paragraph{Recommendation system scaling.} Ardalani et al.~\citep{ardalani2022understanding} show DLRM quality follows power-law-plus-constant with parameter scaling saturating before data scaling. Wukong~\citep{zhang2024wukong} and HSTU~\citep{zhai2024hstu} confirm power-law scaling. These target absolute performance for capacity planning; PS targets the complementary problem of ranking hypotheses relative to a baseline.

\section{Predictable Scaling: Problem Formulation and Framework}
\label{sec:framework}

\subsection{Setting}
\label{sec:setting}

Let $\mathcal{P}_0$ denote a baseline model, and $\{\mathcal{P}_1, \ldots, \mathcal{P}_N\}$ denote $N$ hypothesis configurations. Each yields scalar metric $\ell_i$ (normalized entropy) after convergent training. The \emph{relative performance gap} is:
\begin{equation}
\Delta_i = \frac{\ell_i - \ell_0}{\ell_0}, \quad i \in [N],
\label{eq:gap}
\end{equation}
where $\Delta_i < 0$ indicates improvement. A scaling mechanism $s_j$ produces a more efficient variant with scaled gap $\Delta_i^{(j)} = (\ell_i^{(j)} - \ell_0^{(j)})/\ell_0^{(j)}$. The scaled gaps form observation vector $\mathbf{d}_i = (\Delta_i^{(1)}, \ldots, \Delta_i^{(M)}) \in \mathbb{R}^M$.

\subsection{Scaling Functions}
\label{sec:scaling-functions}

Each scaling mechanism $s_j = (\phi_k, \alpha_l)$ binds a function family to a reduction ratio $\alpha \in (0,1]$. We implement six families (\cref{tab:scaling-functions}); full details in \cref{app:scaling}.

\begin{table}[htbp]
\centering
\begin{tabular}{@{}lllll@{}}
\toprule
\textbf{Fn} & \textbf{Description} & \textbf{Scaled} & \textbf{Dim} & \textbf{$r_k(\alpha)$} \\
\midrule
DD & Uniform subsampling, hash-seeded & Data volume & FLOPs & $\alpha$ \\
HSR & Reduce table rows (min 5K) & Embedding rows & Memory & $\alpha$ \\
ES & Shrink vector width, aligned & Embedding width & FLOPs+Mem & $\alpha$ \\
FS & Prune by importance scores & Input features & FLOPs & $\alpha$ \\
WAS & Scale all dims uniformly & All hidden dims & FLOPs+Mem & $\alpha^2$ \\
DS & Truncate to $\lfloor \alpha L \rfloor$ layers & Layers & FLOPs & $\alpha$ \\
\bottomrule
\end{tabular}
\caption{Scaling functions in PS. Each parameterized by $\alpha \in (0, 1]$.}
\label{tab:scaling-functions}
\end{table}

Multiple functions compose multiplicatively: $r(c) = \prod_{t=1}^{m} r_{k_t}(\alpha_{l_t})$. We define \emph{scaling profiles}---curated compositions producing proxies at a target computational reduction (e.g., $10\times$).

\subsection{Gap Extrapolation}
\label{sec:extrapolation}

A predictor $f_\theta \colon \mathbb{R}^M \to \mathbb{R}$ maps proxy observations to predicted full-scale gaps $\hat{\Delta}_i = f_\theta(\mathbf{d}_i)$. Beyond the raw proxy baseline ($\hat{\Delta}_i = \Delta_i^{(1)}$), we consider:

\paragraph{Linear regression.} $\hat{\Delta}_i = \mathbf{w}^\top \mathbf{d}_i + b$, fit via OLS. Effective when the gap transfer function is approximately affine.

\paragraph{Empirical Gaussian Process (EGP).} A GP whose prior is learned from historical training curves~\citep{lin2026empirical}: the mean function is the pointwise average and the covariance is the empirical sample covariance of historical NE trajectories. The EGP extrapolates NE over \emph{training steps} at a fixed scale point, predicting where the proxy's learning curve will converge. The posterior provides calibrated uncertainty; predictions exceeding a coefficient-of-variation threshold are rejected, signaling that full-scale training is needed.

\subsection{Optimization Objective and Success Criteria}
\label{sec:objectives}

PS replaces full-scale training with efficient proxy runs and extrapolates to the target scale. It jointly optimizes scaling mechanisms $\mathcal{S}$ and predictor $f_\theta$:
\begin{equation}
\hat{\Delta}_i = f_\theta(\mathbf{d}_i), \quad \min_{\mathcal{S}, g \in \mathcal{G}, \theta} \mathcal{E} \;\; \text{s.t.} \;\; |\mathcal{S}| = M,\; \text{compute}(\mathcal{S}) \leq B.
\label{eq:joint}
\end{equation}

The primary metric is MAE: $\mathcal{E} = \frac{1}{N}\sum_{i=1}^N |\hat{\Delta}_i - \Delta_i|$. The secondary metric is Spearman $\rho$ over hypotheses with $|\Delta_i| > \tau_{\text{AA}}$ (the A/A noise floor; formal definition in \cref{app:rank-preservation}).

\paragraph{Regularity conditions.} Four assumptions under which reduced-scale evaluation succeeds: \textbf{(A1)}~Convergence stability---metrics are stable and reproducible; \textbf{(A2)}~Gap transferability---a consistent function $f^*$ maps scaled to full-scale gaps; \textbf{(A3)}~Monotonic signal preservation---scaling preserves gap sign with high probability; \textbf{(A4)}~Low-dimensional predictive structure---$f^*$ admits a low-complexity form. Empirical verification in \cref{app:assumption-verification}.

\subsection{Compute Budget-Optimal Proxy Selection}
\label{sec:proxy-selection}

The selection procedure enumerates feasible configurations satisfying $(N{+}1) \cdot r(c) \cdot C_{\text{full}} \leq B$, trains proxies on a benchmark corpus with known full-scale gaps, and selects $(\mathcal{S}^*, g^*)$ minimizing $\mathcal{E}$ via cross-validation. The enumeration is tractable ($K$ small, coarse ratio grid, depth $m \leq 3$) and pruned via compute-tier filtering. Full procedure in \cref{alg:joint-selection} (\cref{app:proxy-selection}).

\section{Experiments and Results}
\label{sec:results}

We evaluate PS on large-scale industrial datasets. A reproducible public evaluation on Criteo~\citep{criteo2014} with four public models is provided in \cref{app:ps-on-public-data}.

\paragraph{Models.} Two ranking models of increasing complexity (\cref{tab:models}): \textbf{TS-1}, a task-specific feed ranking model; \textbf{FM-1}, a foundation model (${\approx}35\times$ TS-1 FLOPs). Both are multi-task models jointly predicting \emph{CTR} and \emph{CVR} across recommendation surfaces. Both train on $\mathcal{O}(10^{10})$--$\mathcal{O}(10^{11})$ user interaction logs with ${\sim}1500$ dense and ${\sim}700$ sparse, high-cardinality categorical features.

\begin{table*}[t]
\centering
\resizebox{\textwidth}{!}{%
\begin{tabular}{@{}llccccl@{}}
\toprule
\textbf{Model} & \textbf{Primary Task(s)} & \textbf{FLOPs/fwd (M)} & \textbf{Trainers} & \textbf{Samples (B)} & \textbf{Hardware} & \textbf{Proxy Budget Constraint} \\
\midrule
TS-1 (Task-Specific) & Feed CTR + CVR & $3{,}403$ & $112$ & $80$ & NVIDIA H100 & $\Delta$Trainers $\leq -50\%$ \\
FM-1 (Foundation) & Multi-task CTR + CVR & $117{,}734$ & $1{,}024$ & $122$ & NVIDIA H100 & $\Delta$Samples $\leq -50\%$ \\
\bottomrule
\end{tabular}
}
\caption{Summary of the two evaluation models. \emph{FLOPs} denote the Floating-point Operations per forward pass and \emph{QPS} denotes the queries per second for each model type.}
\label{tab:models}
\end{table*}

\paragraph{Setup.} We use \emph{Normalized Entropy} (NE)~\citep{he2014practical} as the evaluation metric ($\text{NE}{<}1$ indicates improvement over base rate). All evaluations use $M{=}1$ proxy. TS-1 evaluates 14 held-out hypotheses across 3 predictors; FM-1 evaluates 7 held-out hypotheses comparing raw proxy and GP. MAE values are normalized by per-model $\tau_{\text{AA}}$.

\begin{figure*}[t]
\centering
\begin{subfigure}[b]{0.32\textwidth}
\centering
\includegraphics[width=\textwidth]{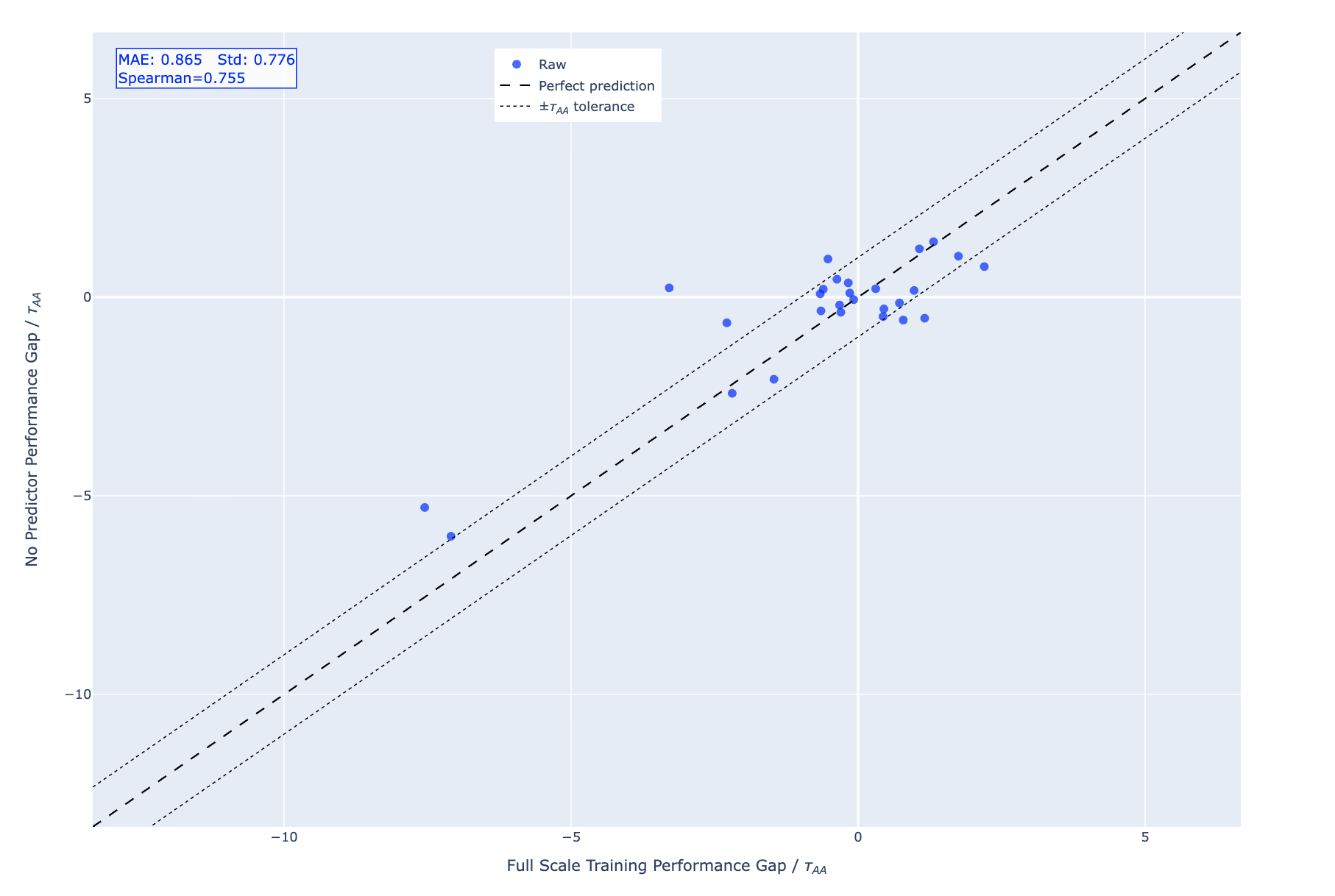}
\caption{Raw proxy}
\end{subfigure}
\hfill
\begin{subfigure}[b]{0.32\textwidth}
\centering
\includegraphics[width=\textwidth]{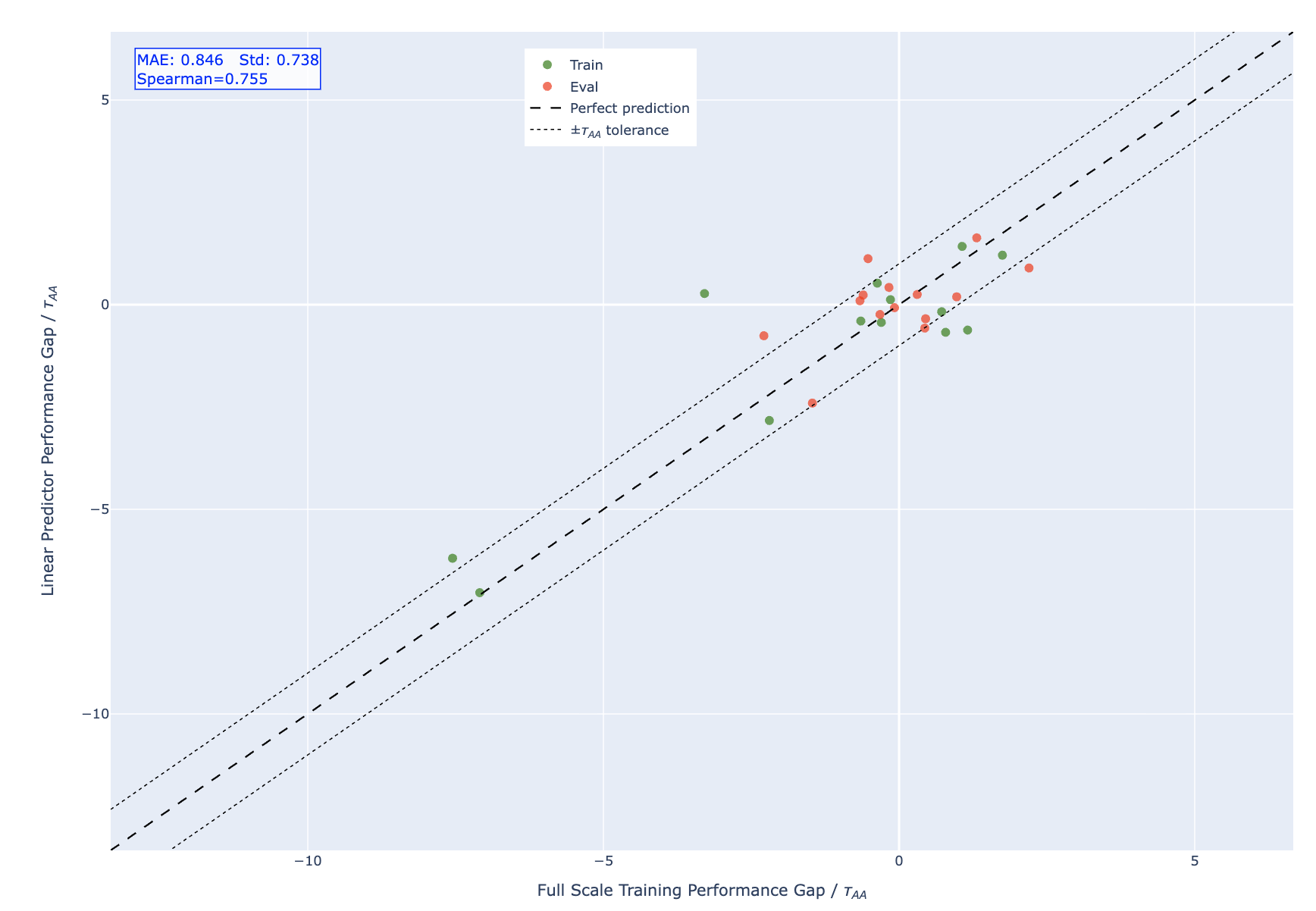}
\caption{Linear}
\end{subfigure}
\hfill
\begin{subfigure}[b]{0.32\textwidth}
\centering
\includegraphics[width=\textwidth]{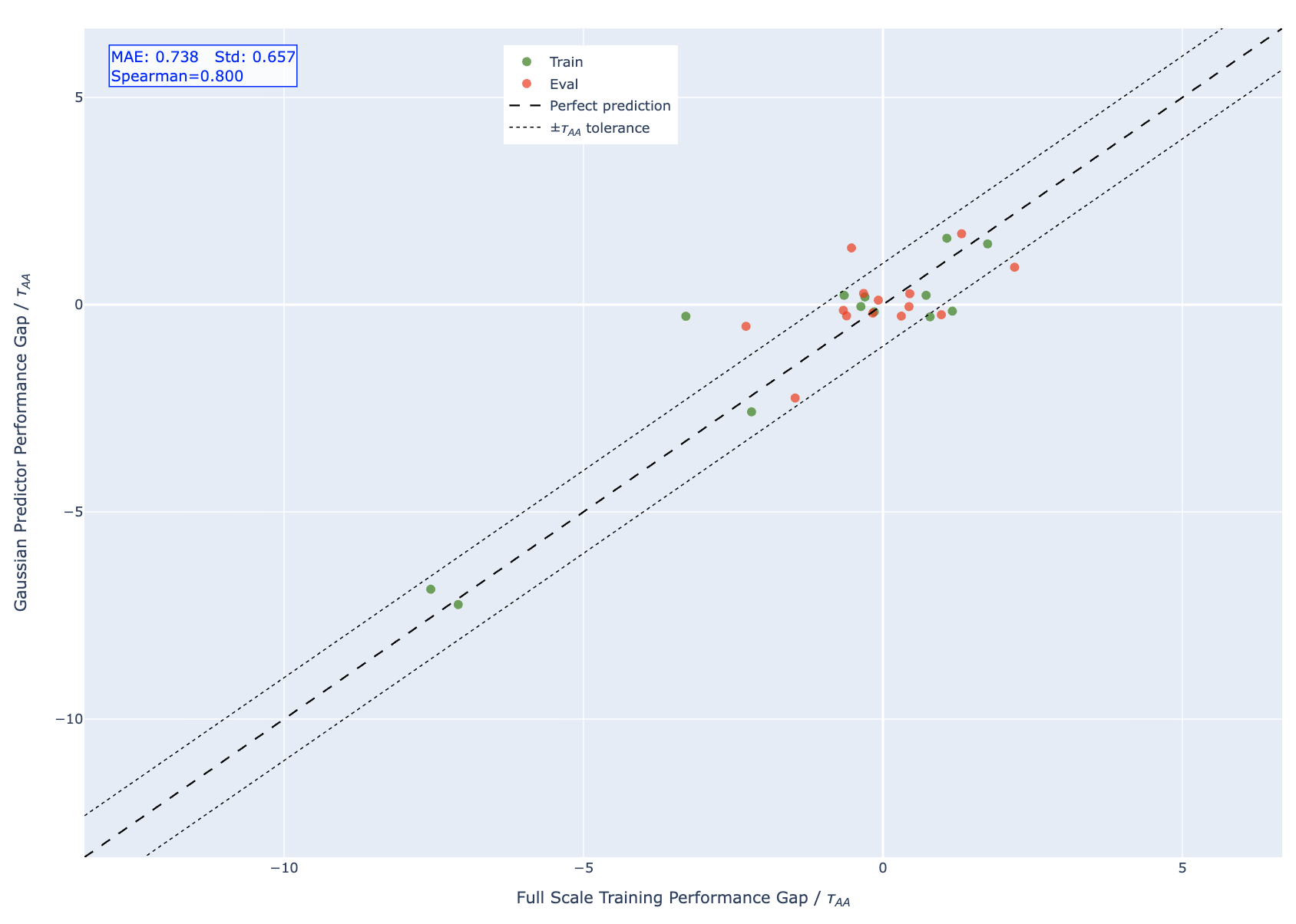}
\caption{GP (EGP)}
\end{subfigure}
\caption{TS-1 predictive fidelity (CTR). Each point is a hypothesis; diagonal = perfect prediction; shaded band = $\pm\tau_{\text{AA}}$. Filled: held-out; open: training.}
\label{fig:ts1-fidelity-comparison}
\end{figure*}

\paragraph{TS-1: Predictor ablation.} \Cref{tab:ts1-accuracy} compares three strategies on TS-1 (CTR head). The raw proxy achieves $0.865\,\tau_{\text{AA}}$ MAE ($\rho = 0.755$); the EGP achieves the best at $0.738\,\tau_{\text{AA}}$ ($\rho = 0.8$)---a $15\%$ relative MAE improvement. Among the 6 actionable hypotheses ($|\Delta| > \tau_{\text{AA}}$), PS achieves $\rho = 1.0$ under all predictors. \Cref{fig:ts1-fidelity-comparison} visualizes this progression.

\begin{table}[t]
\centering
\begin{tabular}{@{}lcrrcr@{}}
\toprule
\textbf{Predictor} & \textbf{MAE ($\times\tau_{\text{AA}}$)} & \textbf{SE} & \textbf{$\Delta$MAE} & \textbf{$\rho$} & \textbf{$\Delta\rho$} \\
\midrule
Raw proxy & $0.865$ & $\pm0.776$ & --- & $0.755$ & --- \\
Linear & $0.846$ & $\pm0.738$ & $-4\%$ & $0.755$ & $0\%$ \\
GP (EGP) & $\bm{0.738}$ & $\pm0.657$ & $-15\%$ & $\bm{0.800}$ & $+6\%$ \\
\bottomrule
\end{tabular}
\caption{TS-1 prediction accuracy (CTR head, 14 held-out hypotheses).}
\label{tab:ts1-accuracy}
\end{table}

\paragraph{FM-1: GP on a foundation model.} \Cref{tab:fm1-accuracy} compares raw proxy and GP on FM-1 across 7 held-out hypotheses. The raw proxy exceeds the noise floor on CTR heads ($2.15\,\tau_{\text{AA}}$, $1.75\,\tau_{\text{AA}}$). The GP reduces these to $0.70$--$0.85\,\tau_{\text{AA}}$---a $57$--$67\%$ MAE improvement---confirming that curve fitting provides substantial value on foundation models where the raw proxy alone is insufficient.

\begin{table}[!htb]
\centering
\begin{tabular}{@{}llcrrcr@{}}
\toprule
\textbf{Head} & \textbf{Predictor} & \textbf{MAE ($\times\tau_{\text{AA}}$)} & \textbf{SE} & \textbf{$\Delta$MAE} & \textbf{$\rho$} & \textbf{$\Delta\rho$} \\
\midrule
\multirow{2}{*}{CTR-1} & Raw proxy & $2.150$ & $\pm1.692$ & --- & $0.607$ & --- \\
 & GP (EGP) & $\bm{0.701}$ & $\pm0.552$ & $-67\%$ & $\bm{0.714}$ & $+18\%$ \\
\midrule
\multirow{2}{*}{CTR-2} & Raw proxy & $1.750$ & $\pm1.466$ & --- & $0.643$ & --- \\
 & GP (EGP) & $\bm{0.753}$ & $\pm0.631$ & $-57\%$ & $\bm{0.714}$ & $+11\%$ \\
\midrule
\multirow{2}{*}{CVR-1} & Raw proxy & $0.950$ & $\pm0.556$ & --- & $0.714$ & --- \\
 & GP (EGP) & $\bm{0.854}$ & $\pm0.500$ & $-10\%$ & $0.714$ & $0\%$ \\
\bottomrule
\end{tabular}
\caption{FM-1 prediction accuracy (7 held-out hypotheses). $\Delta$ columns: change vs.\ raw proxy.}
\label{tab:fm1-accuracy}
\end{table}

\paragraph{Compute reduction.} TS-1 and FM-1 reduce trainers by $86\%$/$88\%$, and samples by $80\%$/$50\%$. Overall, PS achieves $79$--$88\%$ compute reduction across both models (\cref{tab:resource-savings}, \cref{app:compute-reduction-extended}).

\section{Discussion and Conclusion}
\label{sec:discussion}

PS performs best with fast NE convergence, low AA variance, and effect sizes exceeding $\tau_{\text{AA}}$; when these hold, even the raw proxy suffices for triage. The GP's benefit increases with complexity ($15\%$ on TS-1 vs.\ $57$--$67\%$ on FM-1), suggesting curve fitting becomes essential as proxy fidelity degrades at scale. Design trade-offs include scaling aggressiveness (WAS reduces NE gap by $38\%$ relative to ES at $\alpha{=}0.5$; \cref{app:es-vs-was}) and composition risk---sign reversals are detectable on a benchmark corpus and should be rejected above 15--20\% rates. Unlike classical scaling laws~\citep{kaplan2020scaling, hoffmann2022training} predicting absolute performance for homogeneous architectures, PS predicts relative gaps across coupled axes in heterogeneous models, extending to LLMs where DD maps to token subsampling and WAS/DS to width/depth scaling~\citep{tay2022scalinglaws}, reducing per-hypothesis compute and carbon footprint by ${\sim}5$--$8\times$.

Across both evaluation models, PS achieves $79$--$88\%$ compute reduction with noise-floor prediction accuracy and perfect rank preservation $(\rho=1.0)$ among actionable hypotheses. Future directions include multi-scale configurations ($M > 1$), extending to larger foundation models, adapting PS to online metrics beyond NE, and architecture-aware scaling ratio selection (motivated by DCN-v2 findings in \cref{app:public-results}). For newer architectures where regularity conditions are unverified, hybrid strategies combining PS screening with selective full-scale validation offer the best quality balance.

\section*{Acknowledgments}
\label{sec:acknowledgement}

This paper was partially written with the assistance of large language models (LLMs) for text drafting, editing, and structural refinement. All technical content, experimental design, results, and scientific claims were produced and verified by the authors.

\clearpage
\newpage

\bibliographystyle{assets/plainnat}
\bibliography{paper}

\clearpage
\newpage

\beginappendix

\section{Additional Experimental Results}\label{app:results}

\subsection{Hypothesis Categories}\label{app:hypothesis-categories}

In large-scale recommendation systems, model development proceeds through \emph{hypotheses}---atomic configuration changes applied to a baseline that target a specific modeling dimension. Each hypothesis modifies exactly one aspect of the model and falls into one of three categories:

\begin{enumerate}
\item \textbf{Architecture hypotheses} alter the network topology or training procedure: layer fusion, precision changes, distillation loss modifications, attention mechanism changes, activation function swaps, normalization layer modifications, or residual connection restructuring.
\item \textbf{Feature hypotheses} add, remove, or transform input features: deprecation of redundant sparse features, embedding table restructuring, feature crossing modifications, addition of new user or item signals, or changes to feature preprocessing pipelines.
\item \textbf{Data hypotheses} modify the training data pipeline: sampling strategies, deduplication filters, data augmentation schemes, label smoothing, temporal windowing changes, or modifications to negative sampling ratios.
\end{enumerate}

\noindent Evaluating hypotheses at full scale is the primary computational bottleneck in the model development cycle: each candidate change requires a complete training run to assess its impact on offline metrics. PS aims to replace these expensive full-scale runs with efficient proxy evaluations while preserving ranking fidelity among hypotheses.

\subsection{Scaling Configurations}\label{app:scaling-configs}

\Cref{tab:scaling-configs} summarizes the scaling function compositions selected for each model via the compute-optimal proxy selection procedure (\cref{sec:proxy-selection}).

\begin{table}[htbp]
\centering
\begin{tabular}{@{}llllll@{}}
\toprule
\textbf{Model} & \textbf{DD} & \textbf{HSR} & \textbf{ES} & \textbf{Other} & \textbf{Constraint} \\
\midrule
TS-1 & $\alpha{=}0.317$ & $\alpha{=}0.5$ & $\alpha{=}0.25$ & --- & $\Delta$Trainers $\leq -50\%$ \\
FM-1 & $\alpha{=}0.5$ & $\alpha{=}0.2$ & $\alpha{=}0.2$ & FS $0.2$, DS $0.5$ & $\Delta$Samples $\leq -50\%$ \\
\bottomrule
\end{tabular}
\caption{Scaling configurations per model.}
\label{tab:scaling-configs}
\end{table}

\noindent For TS-1, the composition of DD, HSR, and ES reflects the infrastructure constraint of halving the trainer count while preserving data distribution fidelity. For FM-1, aggressive scaling across five axes (DD, HSR, FS, ES, DS) achieves $79\%$ compute reduction under a $50\%$ data-reduction ceiling.

\subsection{Extended Per-Head and Predictor Analysis}\label{app:per-head-extended}

On TS-1, all three prediction strategies fall within $\tau_{\text{AA}}$, and the absolute improvement from raw proxy to EGP is modest ($0.13\,\tau_{\text{AA}}$ in mean, $0.12\,\tau_{\text{AA}}$ in SE). Nonetheless, curve fitting adds meaningful value: the EGP improves Spearman $\rho$ from $0.755$ to $0.8$, tightening the ranking among near-neutral hypotheses, while simultaneously reducing prediction variance across the held-out pool. More importantly, the EGP's primary advantage lies in its calibrated uncertainty estimates (\cref{sec:extrapolation}), which enable automatic rejection of unreliable predictions---a capability absent from both the raw proxy and the linear predictor.

On FM-1, \cref{tab:fm1-accuracy} reveals a striking contrast between prediction heads. The raw proxy is severely degraded on CTR heads (CTR-1: $2.15\,\tau_{\text{AA}}$, CTR-2: $1.75\,\tau_{\text{AA}}$) but already near the noise floor on CVR-1 ($0.95\,\tau_{\text{AA}}$). The GP dramatically improves the CTR heads---reducing MAE by $67\%$ and $57\%$ respectively---while providing only a modest $10\%$ improvement on CVR-1. This asymmetry suggests that the FM-1 scaling configuration (DD + HSR + FS + ES + DS; \cref{app:scaling-configs}) introduces greater distortion for CTR-related representations than for CVR, possibly because CTR predictions rely more heavily on fine-grained embedding interactions that are disrupted by aggressive scaling across five axes. The GP's empirical prior, learned from 9 training hypotheses, successfully corrects this distortion. The identical Spearman $\rho = 0.714$ across all three GP-predicted heads reflects the same 7 held-out hypotheses being ranked in the same order regardless of prediction objective, indicating stable ranking structure. On the raw proxy, rank preservation is lower on CTR heads ($\rho = 0.607$--$0.643$) than on CVR-1 ($\rho = 0.714$), mirroring the MAE pattern.

Confirming these patterns and extending the predictor comparison with a larger foundation model hypothesis pool remain important directions for future work.

\subsection{Compute reduction}\label{app:compute-reduction-extended}

\begin{equation}
\text{Compute Reduction}(\theta_{\text{full}}, \theta_{\text{proxy}}) = 1 - \frac{(1 + \Delta_{\text{Trainers}}) \cdot (1 + \Delta_{\text{Samples}})}{1 + \Delta_{\text{QPS}}}\,,
\label{eq:compute-reduction}
\end{equation}

where $\theta_{\text{full}}$ and $\theta_{\text{proxy}}$ denote the full-scale and proxy configurations, respectively, and $\Delta_{\text{QPS}}$, $\Delta_{\text{Trainers}}$, and $\Delta_{\text{Samples}}$ are the relative percentage changes in throughput or Queries per Sample (QPS), trainer count, and sample volume between the two configurations.

This formula models total training compute as proportional to $\text{trainers} \times \text{samples} / \text{throughput}$: training computation scales with parallelism (trainer count) and data volume (samples), and inversely with per-device throughput (QPS). A decrease in QPS increases per-sample processing time, partially offsetting reductions in other axes. Since PS proxy runs are executed on the same hardware type as their corresponding full-scale baselines (\cref{tab:models}), the relative change in trainer count between proxy and full-scale configurations directly reflects the reduction in compute infrastructure, without confounding differences in per-device throughput.

\begin{table}[htbp]
\centering
\begin{tabular}{@{}lcccc@{}}
\toprule
\textbf{Model} & \textbf{$\Delta$QPS} & \textbf{$\Delta$Samples} & \textbf{$\Delta$Trainers} & \textbf{Compute Red.} \\
\midrule
TS-1 & $-76\%$ & $-80\%$ & $-86\%$ & $88\%$ \\
FM-1 & $-70\%$ & $-50\%$ & $-88\%$ & $79\%$ \\
\bottomrule
\end{tabular}
\caption{Resource efficiency of selected proxy configurations.}
\label{tab:resource-savings}
\end{table}

\section{Formal Definitions}\label{app:formal-definitions}

\subsection{Rank Preservation}\label{app:rank-preservation}

We define the \emph{non-neutral} hypothesis set as
\begin{equation}
\mathcal{I} = \left\{\, i \in [N] \;\middle|\; |\Delta_i| > \tau_{\text{AA}} \,\right\},
\label{eq:non_neutral}
\end{equation}
and compute rank correlation only over this set:
\begin{equation}
\rho_{\text{rank}} = \text{Spearman}\!\left(\text{rank}(\hat{\boldsymbol{\Delta}}_{\mathcal{I}}),\; \text{rank}(\boldsymbol{\Delta}_{\mathcal{I}})\right),
\label{eq:rank_corr}
\end{equation}
where $\hat{\boldsymbol{\Delta}}_{\mathcal{I}}$ and $\boldsymbol{\Delta}_{\mathcal{I}}$ denote the restriction of the predicted and ground-truth gap vectors to the indices in $\mathcal{I}$. Hypotheses with $|\Delta_i| \leq \tau_{\text{AA}}$ are excluded because their gaps are indistinguishable from noise---any ranking among them is arbitrary, and including them would dilute the correlation signal with uninformative ties. A perfect system achieves $\rho_{\text{rank}} = 1$. We note that low proxy error generally entails high rank correlation, but the converse does not hold: a predictor may preserve rankings while systematically distorting gap magnitudes.

\subsection{Normalized Entropy (NE)}\label{app:normalized-entropy}

NE quantifies the predictive quality of a model relative to the empirical base-rate entropy:
\begin{equation}
\text{NE} \;=\; \frac{\mathcal{H}(\hat{\mathbf{y}},\, \mathbf{y})}{\mathcal{H}(p)} \;=\; \frac{-\frac{1}{n}\sum_{k=1}^{n}\bigl[y_k \log \hat{y}_k + (1 - y_k)\log(1 - \hat{y}_k)\bigr]}{-\bigl[p\log p + (1 - p)\log(1 - p)\bigr]}\,,
\label{eq:ne}
\end{equation}
where $y_k \in \{0,1\}$ is the binary label, $\hat{y}_k \in (0,1)$ is the predicted probability, $p = \frac{1}{n}\sum_{k} y_k$ is the marginal positive rate, and $n$ is the number of evaluation examples. A model that perfectly matches the base rate achieves $\text{NE} = 1$; lower values indicate superior predictive performance. NE is preferred over raw log-loss (binary cross-entropy) in industrial settings because it normalizes for variation in base rates across tasks, time periods, and data splits, thereby enabling meaningful comparison across heterogeneous model types and prediction heads. For multi-task models, we report NE independently for each prediction head to capture per-objective scaling behavior.

\section{Scaling Function Details}\label{app:scaling}

This appendix provides implementation details for the scaling functions summarized in \cref{tab:scaling-functions}.

\subsection{Quantized Scaling Operator}\label{app:qscale}

Many scaling functions reduce integer-valued parameters (embedding dimensions, hash sizes, layer widths). Naively multiplying by $\alpha$ and rounding can produce values that are misaligned with hardware (e.g., not multiples of 8 for tensor cores) or degenerate (e.g., embedding dimension of 0). We use a \emph{quantized scaling operator} throughout:

\begin{equation}
\mathrm{QScale}(v, \alpha, q, \tau) = \begin{cases} v & \text{if } v < \tau, \\ q \cdot \lceil \max(\lfloor \alpha \cdot v \rfloor, 1) / q \rceil & \text{otherwise}, \end{cases}
\label{eq:guard-scaling}
\end{equation}

where $v$ is the original value, $q$ is the alignment granularity (e.g., 8), and $\tau$ is a minimum threshold below which values are left unchanged. This ensures scaled values are always positive, hardware-aligned integers.

\subsection{Scaling Function Descriptions}\label{app:scaling-descriptions}

\textbf{Data Downsampling (DD)} is the simplest scaling function: train on a uniformly sampled $\alpha$-fraction of the data using a fixed hash seed for determinism. If the configuration already has a downsampling rate $r_0$, the effective rate becomes $r_0 \cdot \alpha$. Compute reduction is approximately linear in $\alpha$, and because the data distribution is preserved, this is the most reliable scaling function for order preservation.

\textbf{Hash-Size Reduction (HSR)} targets the sparse embedding tables that dominate memory in recommendation models. Each categorical feature is mapped to an embedding table with $h_f$ rows; hash-size reduction scales these to $h_f' = \mathrm{QScale}(h_f, \alpha, 1, 5000)$---features with hash sizes below 5,000 are left unchanged, as further reduction degrades feature resolution. After scaling, all affected tables use explicit hashing to ensure consistent collision behavior. This primarily reduces memory footprint and enables fitting the model on fewer GPUs.

\textbf{Embedding Scaling (ES)} is complementary to HSR: instead of reducing the number of rows, it reduces the \emph{width} of each embedding vector. Each embedding dimension is scaled as $d_i' = \mathrm{QScale}(d_i, \alpha, q_i, 10)$. The alignment factor $q_i$ must simultaneously satisfy memory alignment restrictions ($d_i' \bmod 4 = 0$), attention head compatibility ($d_{\text{model}} \bmod n_{\text{heads}} = 0$), and quantization-specific requirements. We compute $q_i$ as the least common multiple of these individual constraints. Embeddings using FP16 quantization are excluded from scaling entirely.

\paragraph{Compression invariant.} In our architectures, embedding dimension and a compression factor satisfy $d_{\text{emb}} \times k_{\text{compress}} = d_{\text{model}}$. After scaling, we must find a new $(d_{\text{emb}}', k_{\text{compress}}')$ pair satisfying this invariant with $d_{\text{emb}}' \bmod 8 = 0$. We enumerate all valid pairs and select the one minimizing a weighted deviation that prioritizes preserving embedding dimension (which has greater impact on quality) over compression factor:

\begin{equation}
\min_{d' \cdot k' = d_{\text{model}}'} \; 2 \cdot \left|\frac{d' - d_{\text{emb}}}{d_{\text{emb}}}\right| + \left|\frac{k' - k_{\text{compress}}}{k_{\text{compress}}}\right|.
\end{equation}

\textbf{Feature Scaling (FS)} prunes low-importance features using pre-computed feature importance scores, retaining the top $\lfloor \alpha \cdot F \rfloor$ features. Two design choices handle the complexity of feature sets:

\emph{Unified embedding groups.} Features sharing a single embedding table are treated atomically---the group is kept or removed as a whole, based on its average importance, to avoid corrupting shared table structures.

\emph{Dependency resolution.} Features derived from other features via hash transformations create dependency chains. When a feature is removed, all its dependents must also be removed. We resolve this via breadth-first traversal from the removed set. Sparse and dense features are scaled independently.

\textbf{Whole Architecture Scaling (WAS)} reduces \emph{all} hidden dimensions simultaneously---embedding widths, dense layers, specialized sub-networks, task-specific heads, cross-network ranks, attention dimensions, and transformer token counts---by the same ratio $\alpha$. Each architecture parameter $p_j$ is scaled to $\max(\lceil \alpha \cdot p_j \rceil_{\text{align}}, p_j^{\min})$, rounding up to the nearest multiple of the alignment (typically 8) and enforcing a minimum value (typically 1). Because every dimension shrinks proportionally, compute reduction is approximately quadratic in $\alpha$ (i.e., $r_k(\alpha) = \alpha^2$).

The challenge is that architecture parameters are not independent. They are connected by structural constraints:

\begin{itemize}
\item $d_{\text{emb}} \times k_{\text{compress}} = d_{\text{model}}$ (embedding-to-model dimension);
\item $d_{\text{model}} \bmod n_{\text{heads}} = 0$ (attention head divisibility);
\item Expert output dimensions must sum to a target (MoE layers).
\end{itemize}

These constraints form a dependency graph. A dependency resolver scales independent parameters first via topological ordering, then adjusts dependent ones to satisfy constraints.

\textbf{Depth Scaling (DS)} reduces the number of transformer layers by truncating the layout to the first $\lfloor \alpha \cdot L \rfloor$ layers (minimum 1). The deepest layers are removed first, and all dependent configurations---prefetch schedules, expert-parallelism assignments, per-layer compressed embeddings, residual connections to removed layers, and per-layer auxiliary outputs---are pruned. In practice, we implement this as architecture-specific variants, one per model family, each handling its particular dependency structure. WAS and DS are orthogonal and can be composed.

\section{Assumption Verification}
\label{app:assumption-verification}

Assumptions \textbf{(A1)}--\textbf{(A3)} from \cref{sec:framework} can be partially verified empirically for any given model family and scaling mechanism.

\paragraph{(A1) Convergence stability.} We verify convergence by monitoring the NE trajectory of each training run: all runs (both proxy and full-scale) exhibit monotonically decreasing NE that plateaus within the final $10\%$ of training steps. The standard deviation of terminal NE across repeated baseline runs defines $\tau_{\text{AA}}$, which is small relative to the hypothesis gaps of interest.

\paragraph{(A2) Gap transferability.} The scatter plots in \cref{fig:ts1-fidelity-comparison} provide visual evidence of gap transferability: proxy gaps cluster tightly around the identity line, with residuals uncorrelated with gap magnitude. The low MAE values ($< 1.03\,\tau_{\text{AA}}$ across all models and heads) further support this assumption.

\paragraph{(A3) Monotonic signal preservation.} For each scaling function, we compute the sign-reversal rate: the fraction of hypotheses for which $\text{sign}(\Delta_i^{(j)}) \neq \text{sign}(\Delta_i)$. On TS-1, the selected proxy configuration (DD + HSR + ES) achieves a sign-reversal rate below $10\%$, concentrated among near-neutral hypotheses with $|\Delta_i| < \tau_{\text{AA}}$. When a scaling mechanism exhibits high sign-reversal rates for actionable hypotheses, the framework provides a principled signal to exclude that mechanism from $\mathcal{S}$.

\section{Comparison of Scaling Functions}\label{app:es-vs-was}

We compare two architecture scaling strategies: \textbf{ES}, which applies the scale ratio only to embedding-related dimensions while leaving dense layers, specialized sub-networks, task heads, and cross-network ranks at full capacity; and \textbf{WAS}, which applies the ratio uniformly to all dimension parameters.

\begin{table}[htbp]
\centering
\begin{tabular}{@{}lcc@{}}
\toprule
\textbf{Component} & \textbf{ES} & \textbf{WAS} \\
\midrule
Embedding tables & \checkmark & \checkmark \\
User-history embeddings & \checkmark & \checkmark \\
Interaction $d_\text{model}$ & \checkmark & \checkmark \\
Dense layers & --- & \checkmark \\
Specialized sub-networks & --- & \checkmark \\
Task-specific heads & --- & \checkmark \\
Cross-network rank & --- & \checkmark \\
Attention tokens & --- & \checkmark \\
\bottomrule
\end{tabular}
\caption{Components scaled by ES and WAS.}
\label{tab:es-vs-was-components}
\end{table}

\paragraph{NE fidelity and training efficiency.} \Cref{tab:es-vs-was} compares ES and WAS at scaling ratio $0.5$ on a large foundation model (${\approx}100\times$ TS-1 FLOPs, $1{,}024$ trainers on NVIDIA B200). WAS reduces the NE gap from $-0.0090$ to $-0.0056$ (38\% improvement) while delivering 25\% higher QPS and 42.9\% lower FLOPs per sample.

\begin{table}[htbp]
\centering
\begin{tabular}{@{}lcc c@{}}
\toprule
\textbf{Metric} & \textbf{ES} & \textbf{WAS} & \textbf{$\Delta$} \\
\midrule
NE Gap & $-0.0090$ & $-0.0056$ & $38\%\!\downarrow$ \\
\midrule
QPS (k) & 194.9 & 244.1 & $+25.2\%$ \\
FLOPs/sample (M) & 109{,}357.9 & 62{,}440.4 & $-42.9\%$ \\
FLOPs reduction vs.\ full & 68.4\% & 82.0\% & --- \\
\bottomrule
\end{tabular}
\caption{ES vs.\ WAS at $\alpha{=}0.5$ on Foundation Model. NE gap: signed difference (negative = proxy underperforms).}
\label{tab:es-vs-was}
\end{table}

\paragraph{Analysis.} The fidelity gap reflects \emph{dimensional asymmetry} in tightly coupled architectures. Under ES, embedding-adjacent dimensions shrink while downstream components remain at full width. For a foundation model whose architecture features deep multiplicative interactions between embeddings, specialized sub-networks, and attention mechanisms, this width mismatch disrupts learned feature interactions, inflating the NE gap. WAS eliminates this asymmetry by scaling all components uniformly. We adopt WAS for tightly coupled models (FM-1), and ES for simpler, embedding-dominated models (TS-1).

\section{Compute Budget-Optimal Proxy Selection: Full Procedure}\label{app:proxy-selection}

\paragraph{Design space.} Let $\Phi = \{\phi_1, \ldots, \phi_K\}$ denote the set of $K$ candidate scaling functions (\cref{tab:scaling-functions}), and let $\mathcal{A} = \{\alpha_1, \ldots, \alpha_L\}$ denote a discrete grid of candidate scaling ratios (e.g., $\{0.1, 0.2, \ldots, 0.9\}$). A \emph{scaling configuration} is a tuple $c = ((\phi_{k_1}, \alpha_{l_1}), \ldots, (\phi_{k_m}, \alpha_{l_m}))$ specifying which functions to apply and at what ratios. The effective compute reduction is $r(c) = \prod_{t=1}^{m} r_{k_t}(\alpha_{l_t})$, where each $r_k(\alpha)$ is given in \cref{tab:scaling-functions}. Let $\mathcal{G} = \{g_1, \ldots, g_P\}$ denote the set of candidate prediction models (linear regression and EGP as described in \cref{sec:extrapolation}). The full design space is $\mathcal{C} \times \mathcal{G}$.

\paragraph{Benchmark corpus.} Selection is performed using a \emph{benchmark corpus} $\mathcal{B} = \{(\mathcal{P}_0, \mathcal{P}_1, \ldots, \mathcal{P}_N)\}$ of hypotheses for which ground-truth full-scale gaps $\boldsymbol{\Delta} = (\Delta_1, \ldots, \Delta_N)$ are known. This corpus is accumulated over time from experiments that have been trained to convergence. The complete procedure is given in \cref{alg:joint-selection}.

\begin{algorithm}[t]
\caption{Joint Selection of Scaling Configuration and Prediction Model}
\label{alg:joint-selection}
\begin{algorithmic}[1]
\REQUIRE Scaling function families $\Phi = \{\phi_1, \ldots, \phi_K\}$; ratio grid $\mathcal{A} = \{\alpha_1, \ldots, \alpha_L\}$; prediction model candidates $\mathcal{G} = \{g_1, \ldots, g_P\}$; benchmark corpus $\mathcal{B}$ with ground-truth gaps $\boldsymbol{\Delta}$; budget $B$
\ENSURE Optimal scaling configuration set $\mathcal{S}^*$ and prediction model $g^*$
\STATE Stage 1: Enumerate feasible scaling configurations
\STATE $\mathcal{C} \leftarrow \emptyset$
\FOR{each composition $(\phi_{k_1}, \alpha_{l_1}), \ldots, (\phi_{k_m}, \alpha_{l_m})$ with $m \leq M_{\max}$}
    \STATE $c \leftarrow ((\phi_{k_1}, \alpha_{l_1}), \ldots, (\phi_{k_m}, \alpha_{l_m}))$
    \IF{$(N+1) \cdot r(c) \cdot C_{\text{full}} \leq B$}
        \STATE $\mathcal{C} \leftarrow \mathcal{C} \cup \{c\}$
    \ENDIF
\ENDFOR
\STATE Stage 2: Run scaled experiments on benchmark corpus
\FOR{each $c \in \mathcal{C}$}
    \STATE Apply scaling configuration $c$ to all hypotheses in $\mathcal{B}$
    \STATE Train $\mathcal{P}_0^{(c)}, \mathcal{P}_1^{(c)}, \ldots, \mathcal{P}_N^{(c)}$ to convergence
    \STATE Compute scaled gaps: $\Delta_i^{(c)} = (\ell_i^{(c)} - \ell_0^{(c)}) / \ell_0^{(c)}$ for all $i \in [N]$
\ENDFOR
\STATE Stage 3: Train and evaluate prediction models
\STATE $\text{best\_score} \leftarrow \infty$
\FOR{each subset $\mathcal{S} \subseteq \mathcal{C}$ with $|\mathcal{S}| = M$}
    \STATE Construct observation matrix $\mathbf{D} \in \mathbb{R}^{N \times M}$ from $\mathcal{S}$
    \FOR{each prediction model $g_p \in \mathcal{G}$}
        \STATE Fit $f_\theta \leftarrow g_p(\mathbf{D}, \boldsymbol{\Delta})$ via cross-validation on $\mathcal{B}$
        \STATE Compute predictions: $\hat{\Delta}_i = f_\theta(\mathbf{d}_i)$ for all $i \in [N]$
        \STATE Evaluate: $\mathcal{E} = \frac{1}{N} \sum_{i=1}^{N} |\hat{\Delta}_i - \Delta_i|$
        \IF{$\mathcal{E} < \text{best\_score}$}
            \STATE $\mathcal{S}^* \leftarrow \mathcal{S}$; \quad $g^* \leftarrow g_p$; \quad $\text{best\_score} \leftarrow \mathcal{E}$
        \ENDIF
    \ENDFOR
\ENDFOR
\RETURN $(\mathcal{S}^*, g^*)$
\end{algorithmic}
\end{algorithm}

\paragraph{Practical considerations.} The full enumeration in \cref{alg:joint-selection} is tractable because: (i) the number of scaling function families $K$ is small, (ii) the ratio grid $\mathcal{A}$ is coarse (typically 5--9 values), and (iii) we limit composition depth to $m \leq M_{\max} = 3$. The dominant compute is Stage~2---training the scaled models on the benchmark corpus---which is a one-time offline compute investment. Once the observation matrix $\mathbf{D}$ is computed, Stage~3 (model fitting and evaluation) is computationally negligible. We further reduce the search space via \emph{compute-tier filtering}: we retain only configurations whose compute reduction falls within a target range (e.g., $5\times$--$15\times$), discarding configurations that are either too computationally expensive to provide meaningful savings or too aggressively scaled to retain useful signal.

\section{Curve Fitting Details}\label{app:curve-fitting-details}

\subsection{Reference Curve Selection}\label{app:reference-selection}

Both linear regression and the EGP require selecting appropriate reference curves from the historical library. We apply two filters sequentially:

\textbf{Gap trend filter.} For each reference $r$, we compute the relative NE gap with the target at each time step: $g_t = (\ell_{\text{target}}(t) - \ell_r(t)) / \ell_{\text{target}}(t)$. The \emph{gap trend} is the range of this relative gap (in percentage points): $\mathrm{GapTrend}(r) = 100 \cdot |\max_t g_t - \min_t g_t|$. References exceeding a threshold $\delta_{\text{gap}}$ are excluded---a large range indicates the reference and target have divergent dynamics.

\textbf{Correlation filter.} Among remaining references, we compute Pearson correlation with the target over the observed interval (skipping the first 20 time steps to avoid noisy early-training data). We keep only the top-$K'$ by correlation, requiring $\rho \geq 0.9999$. If no reference meets this threshold, the prediction is rejected.

\textbf{Why $K' = 1$?} We find that using a single reference is more robust than larger bases. With $K' > 1$, OLS can overfit to noise in the reference-target alignment, especially when the observed window is short. The extremely high correlation threshold ($0.9999$) compensates---at this level, a single reference already provides an excellent basis.

\subsection{Empirical Gaussian Process Details}\label{app:gp-details}

\textbf{Preprocessing.} Training progress is min-max normalized to $[0, 1]$. NE values are standardized using the global mean and standard deviation across all reference curves. Predictions are un-standardized before returning.

\textbf{Empirical prior.} The mean function is the pointwise average of $R$ historical curves ($R = 13$ for TS-1, $R = 9$ for FM-1, corresponding to training hypotheses), linearly interpolated to query points. The covariance is the sample covariance:

\begin{equation}
k(x, x') = \frac{1}{R - 1} \sum_{i=1}^{R} \left[y_i(x) - \mu(x)\right] \left[y_i(x') - \mu(x')\right].
\end{equation}

This non-parametric kernel directly captures the correlation patterns in training curves without requiring manual kernel engineering.

\textbf{Rejection criterion.} The mean coefficient of variation $\mathrm{CV} = \mathrm{mean}(\hat{\sigma} / |\hat{\mu}|)$ over the prediction horizon serves as a confidence check. When $\mathrm{CV}$ exceeds a task-dependent threshold, the prediction is deemed unreliable.

\section{Predictable Scaling on Public Data}
\label{app:ps-on-public-data}

To complement the evaluation and support reproducibility, we evaluate PS on four widely-used public recommendation model architectures---DLRM~\citep{naumov2019dlrm}, DeepFM~\citep{guo2017deepfm}, xDeepFM~\citep{lian2018xdeepfm}, and DCN-v2~\citep{wang2021dcnv2}---trained on the Criteo Terabyte dataset~\citep{criteo2014}, a standard large-scale click-through rate prediction benchmark comprising approximately $4$ billion labeled user interaction records with $13$ integer and $26$ categorical features. These four architectures span the dominant paradigms in industrial recommendation: DLRM couples sparse embedding lookups with dense feature interactions via dot products, DeepFM jointly models low-order feature interactions through a factorization machine and high-order interactions through a deep network, xDeepFM extends this with a Compressed Interaction Network (CIN) that learns explicit bounded-degree feature interactions at the vector level, and DCN-v2 introduces explicit cross-network layers that learn bounded-degree feature interactions alongside a deep network. For each model, we construct scaled-down proxies using the scaling functions described in \cref{sec:scaling-functions} and evaluate all three prediction strategies---raw proxy, linear predictor (LP), and Empirical Gaussian Process (EGP)---across six scaling configurations (\cref{tab:public-dlrm-all}--\cref{tab:public-dcnv2-all}). This public evaluation serves two purposes: it validates that PS generalizes beyond full scale model families to architecturally diverse public models, and it provides a fully reproducible experimental setup against which future work can be compared.

\subsection{Hypothesis Design}
\label{app:public-hypotheses}

To evaluate PS on public models, we design a pool of $27$ hyperparameter hypotheses applied to a shared baseline configuration: embedding dimension $d_{\text{emb}} = 16$, hidden layers $[256]$, dropout rate $0.3$, and learning rate $10^{-3}$. The hypotheses span five categories that systematically probe distinct modeling dimensions:

\begin{enumerate}
\item \textbf{Embedding dimension sweep} (7 hypotheses). Vary the sparse embedding dimension $d_{\text{emb}} \in \{2, 4, 8, 16, 32, 64, 128\}$, controlling the representational capacity allocated to each categorical feature while holding all other hyperparameters at baseline. \emph{Category: Feature Hypothesis.}
\item \textbf{DNN width sweep} (7 hypotheses). Vary the hidden layer widths from a single narrow layer ($[32]$) through two-layer ($[64, 32]$, $[128, 64]$), three-layer ($[256, 128, 64]$, $[512, 256, 128]$), and four-layer ($[1024, 512, 256, 128]$, $[1024, 1024, 512, 256]$) configurations, jointly scaling width and depth of the dense interaction network. \emph{Category: Architecture Hypothesis.}
\item \textbf{DNN depth sweep} (5 hypotheses). Fix the per-layer width at $256$ and vary only depth: $[256]^{\times k}$ for $k \in \{1, 2, 4, 6, 8\}$, isolating the effect of network depth from width. \emph{Category: Architecture Hypothesis.}
\item \textbf{Joint embedding--architecture sweep} (4 hypotheses). Co-vary embedding dimension and DNN configuration simultaneously, ranging from a minimal model ($d_{\text{emb}}{=}2$, $[32]$) through intermediate configurations ($d_{\text{emb}}{=}4$, $[64, 32]$; $d_{\text{emb}}{=}64$, $[512, 256, 128]$) to a maximal model ($d_{\text{emb}}{=}128$, $[1024, 512, 256, 128]$), testing how sparse and dense capacity interact when scaled together. \emph{Category: Architecture Hypothesis.}
\item \textbf{Regularization and training} (4 hypotheses). Vary dropout rate $\in \{0.0, 0.5, 0.8\}$ (baseline: $0.3$) and learning rate ($10^{-2}$; baseline: $10^{-3}$), probing sensitivity to training-time hyperparameters that do not alter model architecture. \emph{Category: Architecture Hypothesis: Training Hyperparameters.}
\end{enumerate}

\paragraph{Effect size distribution.} The hypothesis pool exhibits a diverse range of effect sizes: six hypotheses are neutral, nine exhibit weak effects, and twelve show strong effects. This distribution ensures that the PS predictor is evaluated across the full spectrum of hypothesis magnitudes---from subtle, near-threshold changes that stress the predictor's sensitivity, to large-effect changes where correct ranking is most actionable. We categorize hypotheses into three actionability tiers based on NE gap magnitude relative to the empirical noise floor: \emph{neutral} ($|\Delta| < 0.55\%$), \emph{weak} ($0.55$--$1.5\%$), and \emph{strong} ($> 1.5\%$).

\paragraph{Train--test split.} The $27$ hypotheses are partitioned into $17$ training and $10$ held-out evaluation hypotheses via stratified sampling, ensuring each effect category (neutral, weak, strong) is represented in both sets. This prevents the predictor from being evaluated exclusively on easy (strong-signal) or hard (near-neutral) hypotheses.

\subsection{Scaling Configurations}
\label{app:public-scaling-configs}

Unlike the full scale models (TS-1, FM-1), where proxy compute budgets are constrained by infrastructure-specific limits (trainer counts, QPS, sample volumes), the public benchmark operates under a simpler objective: reducing the overall training computation. Since no infrastructure constraints apply, we select multiple scaling configurations at a uniform factor of $0.5$ along different resource axes, enabling a systematic comparison of how each scaling dimension affects proxy fidelity. The six configurations (\cref{tab:public-scaling-configs}) span single-axis and multi-axis scaling:

\begin{table}[htbp]
\centering
\begin{tabular}{@{}llcc@{}}
\toprule
\textbf{Config} & \textbf{HSR} & \textbf{ES} & \textbf{DD} \\
\midrule
hash\_only & $0.5$ & $1.0$ & $1.0$ \\
emb\_only & $1.0$ & $0.5$ & $1.0$ \\
data\_downsample & $1.0$ & $1.0$ & $0.5$ \\
hash\_emb & $0.5$ & $0.5$ & $1.0$ \\
hash\_data & $0.5$ & $1.0$ & $0.5$ \\
all\_scaled & $0.5$ & $0.5$ & $0.5$ \\
\bottomrule
\end{tabular}
\caption{Scaling configurations for public benchmark evaluation. Each configuration scales the indicated dimensions by a factor of $0.5$, targeting overall training compute reduction.}
\label{tab:public-scaling-configs}
\end{table}

\noindent By varying which dimensions are scaled, we isolate the contribution of each resource axis to proxy fidelity, and by including multi-axis compositions (hash\_emb, hash\_data, all\_scaled), we test whether combining scaling dimensions introduces interaction effects that degrade prediction quality. The uniform $0.5$ ratio across all configurations simplifies interpretation and ensures that observed differences in proxy fidelity are attributable to the choice of scaling dimensions rather than differing reduction magnitudes.

\subsection{Cross-Model Results and Analysis}
\label{app:public-results}

\Cref{tab:public-dlrm-all}--\cref{tab:public-dcnv2-all} present prediction accuracy across all four public models, six scaling configurations, and three predictors. \Cref{fig:dlrm-scatter-all}--\cref{fig:dcnv2-scatter-all} show the corresponding predicted-vs.-actual scatter plots. We organize the analysis around the paper's three core claims, identifying which results strengthen each claim and where limitations emerge.

\subsubsection{GP Predictor Value}
\label{app:public-gp-value}

\begin{figure*}[t]
\centering
\begin{subfigure}[b]{0.65\textwidth}
\centering
\includegraphics[width=\textwidth]{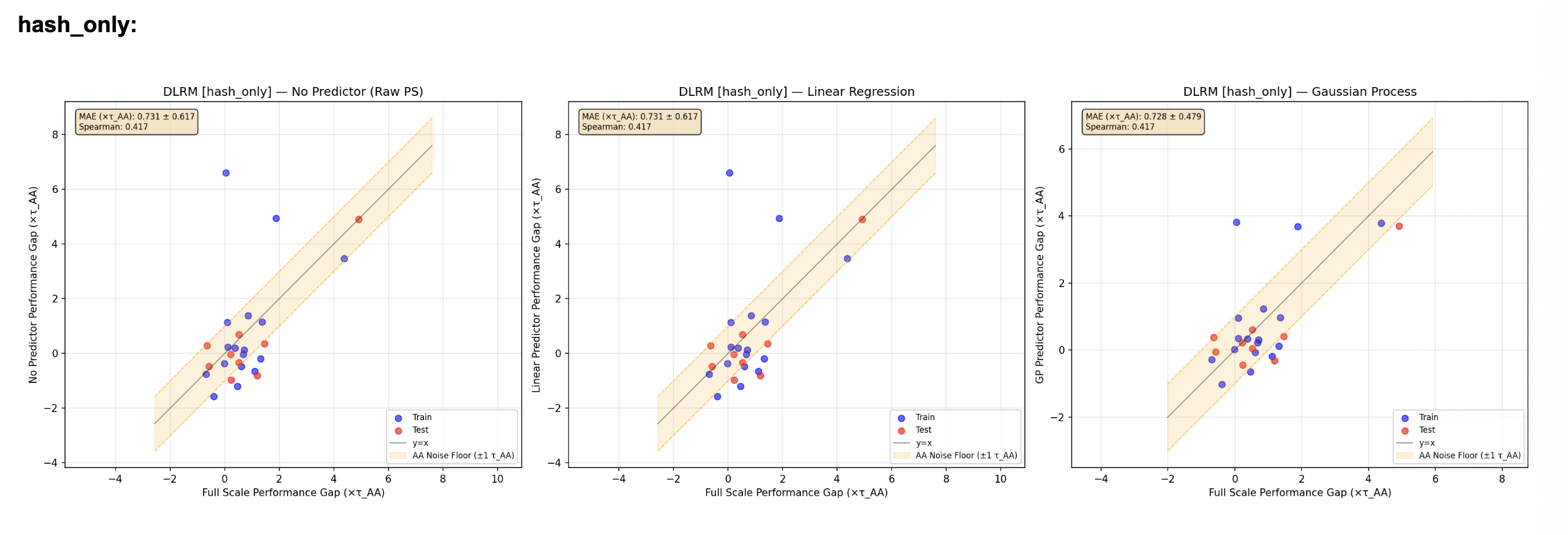}
\caption{Hash-only scaling}
\label{fig:dlrm-hash-only}
\end{subfigure}
\begin{subfigure}[b]{0.65\textwidth}
\centering
\includegraphics[width=\textwidth]{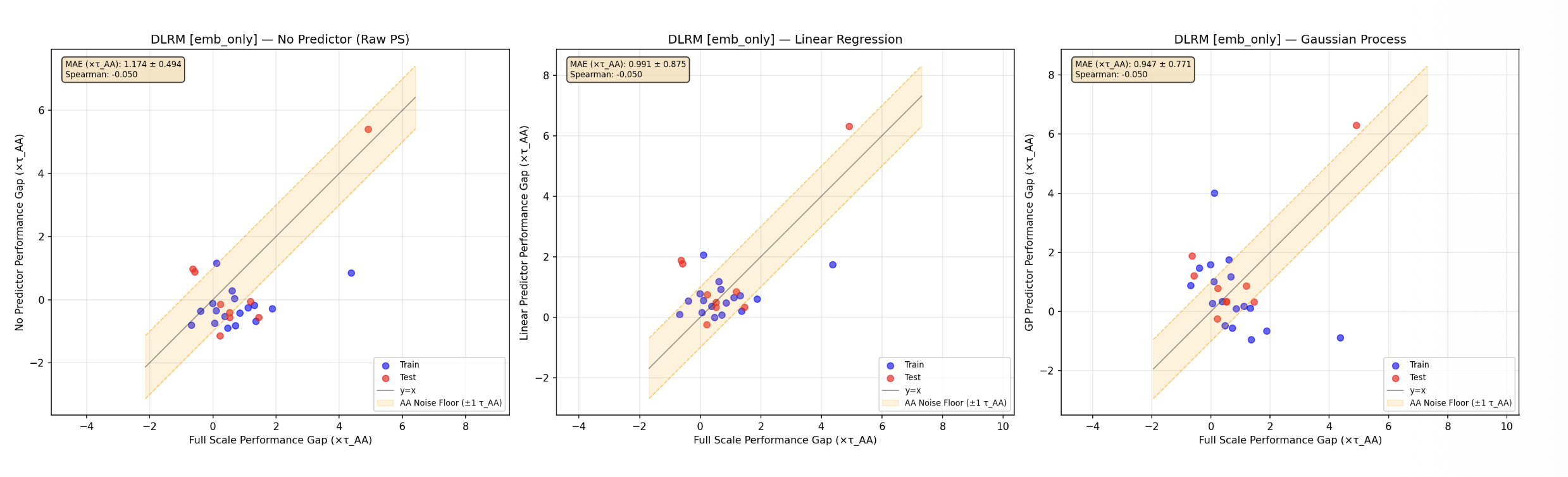}
\caption{Embedding-only scaling}
\label{fig:dlrm-emb-only}
\end{subfigure}
\begin{subfigure}[b]{0.65\textwidth}
\centering
\includegraphics[width=\textwidth]{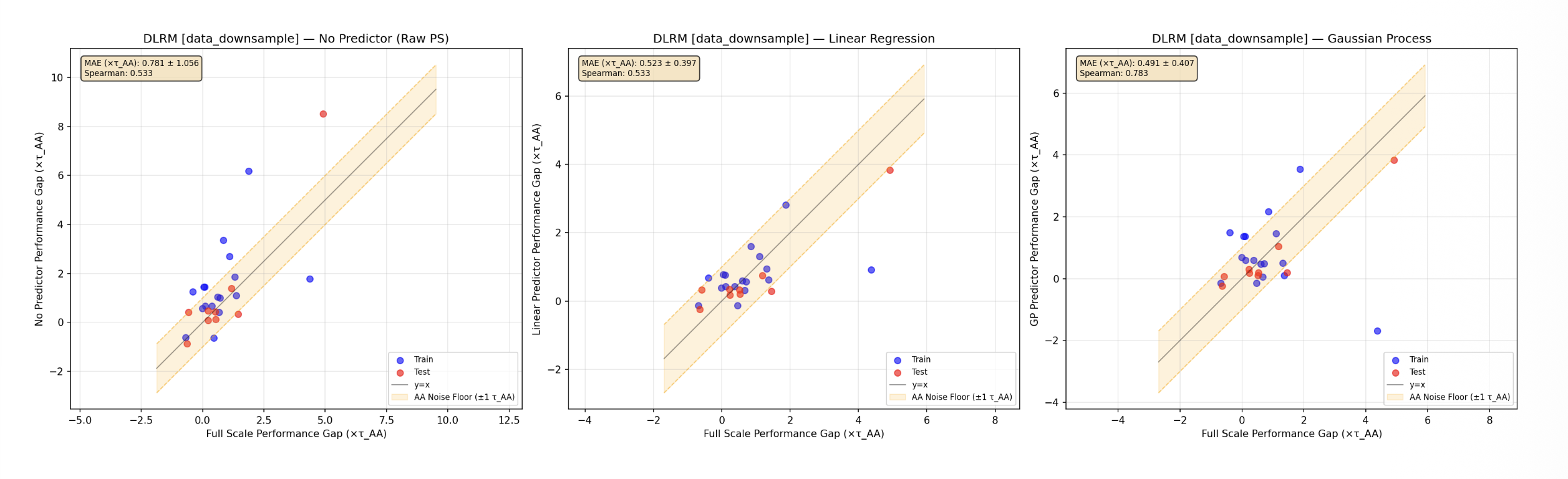}
\caption{Data downsampling}
\label{fig:dlrm-data-ds}
\end{subfigure}
\begin{subfigure}[b]{0.65\textwidth}
\centering
\includegraphics[width=\textwidth]{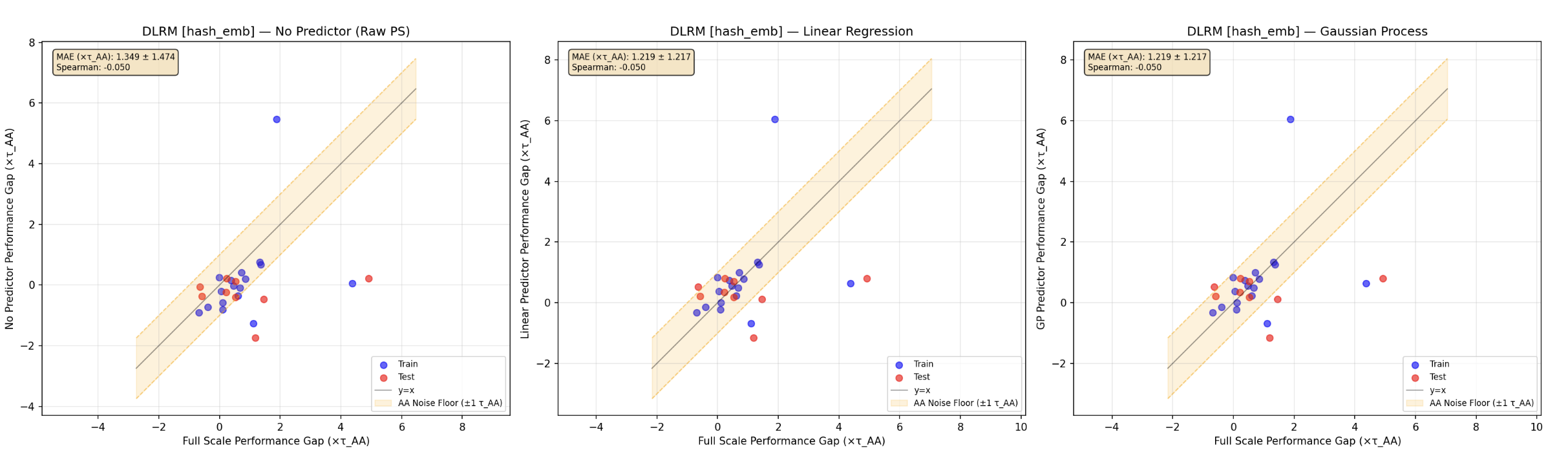}
\caption{Hash + embedding scaling}
\label{fig:dlrm-hash-emb}
\end{subfigure}
\begin{subfigure}[b]{0.65\textwidth}
\centering
\includegraphics[width=\textwidth]{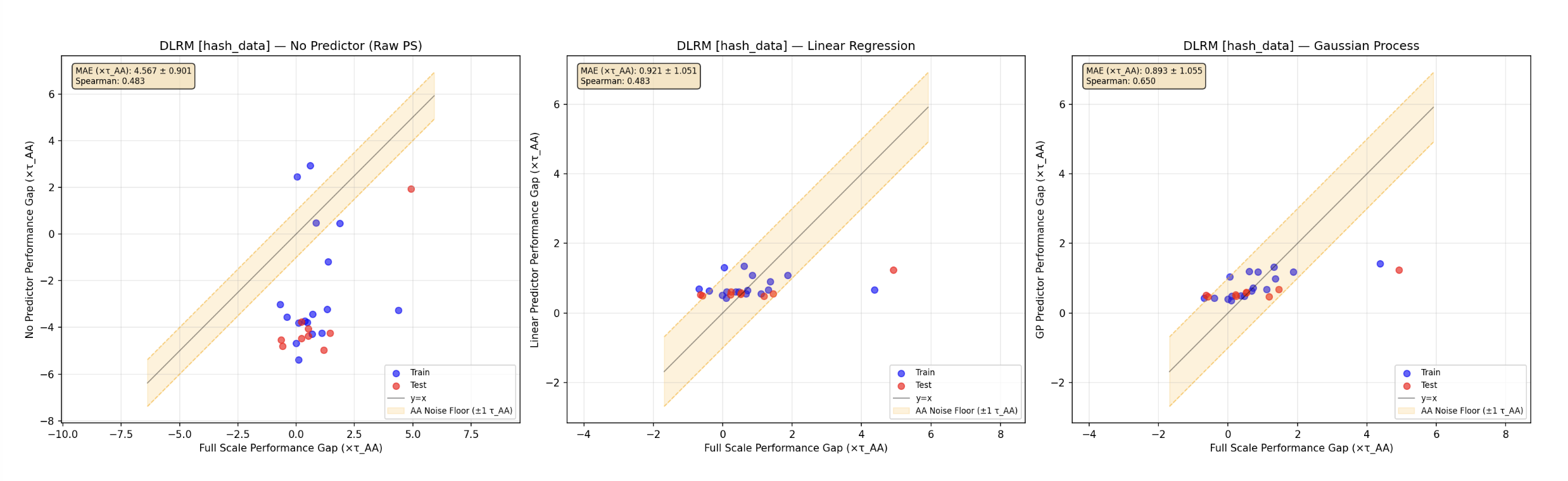}
\caption{Hash + data scaling}
\label{fig:dlrm-hash-data}
\end{subfigure}
\begin{subfigure}[b]{0.65\textwidth}
\centering
\includegraphics[width=\textwidth]{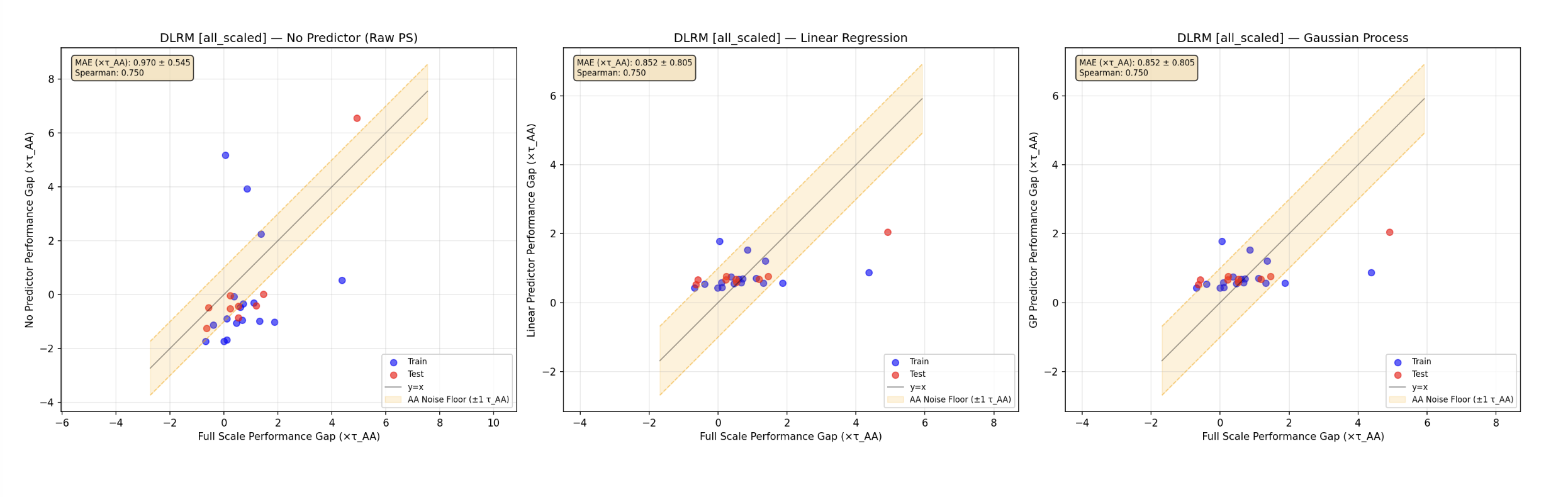}
\caption{All dimensions scaled}
\label{fig:dlrm-all-scaled}
\end{subfigure}
\caption{Predicted vs.\ actual accuracy gap for six scaling configurations on the public DLRM benchmark. Each point represents one hypothesis; the dashed line indicates perfect prediction. Tighter clustering around the diagonal indicates higher proxy fidelity.}
\label{fig:dlrm-scatter-all}
\end{figure*}

\begin{figure*}[t]
\centering
\begin{subfigure}[b]{0.65\textwidth}
\centering
\includegraphics[width=\textwidth]{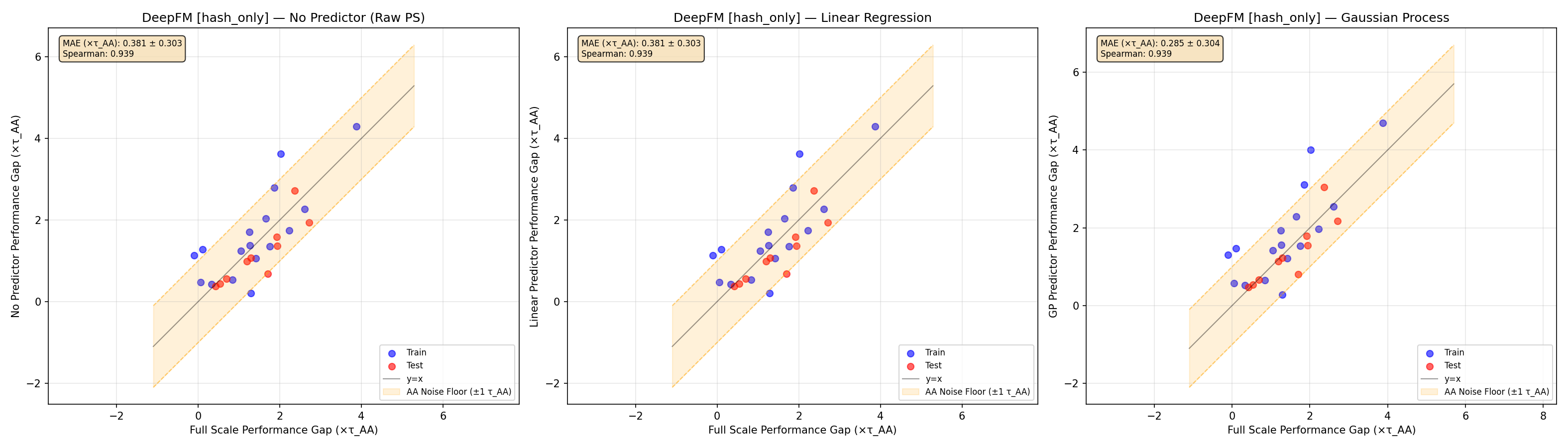}
\caption{Hash-only scaling}
\label{fig:deepfm-hash-only}
\end{subfigure}
\begin{subfigure}[b]{0.65\textwidth}
\centering
\includegraphics[width=\textwidth]{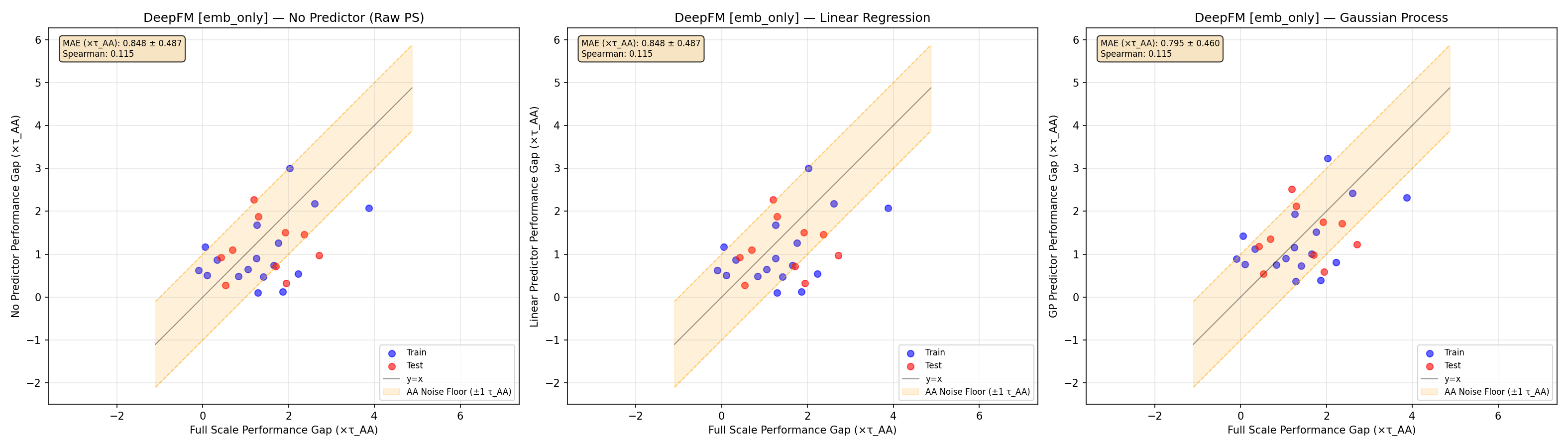}
\caption{Embedding-only scaling}
\label{fig:deepfm-emb-only}
\end{subfigure}
\begin{subfigure}[b]{0.65\textwidth}
\centering
\includegraphics[width=\textwidth]{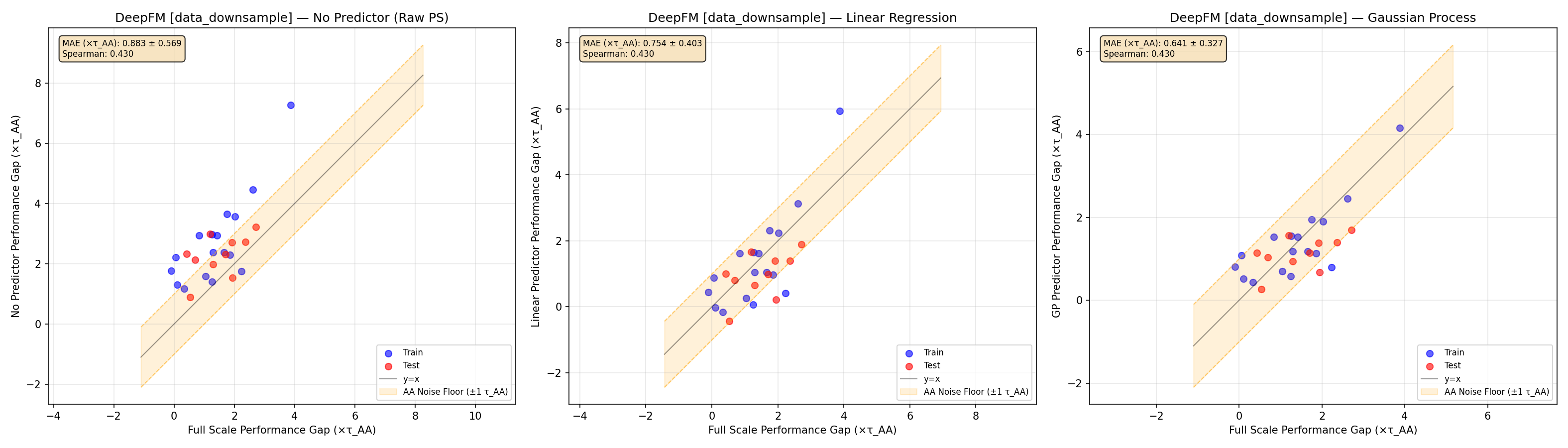}
\caption{Data downsampling}
\label{fig:deepfm-data-ds}
\end{subfigure}
\begin{subfigure}[b]{0.65\textwidth}
\centering
\includegraphics[width=\textwidth]{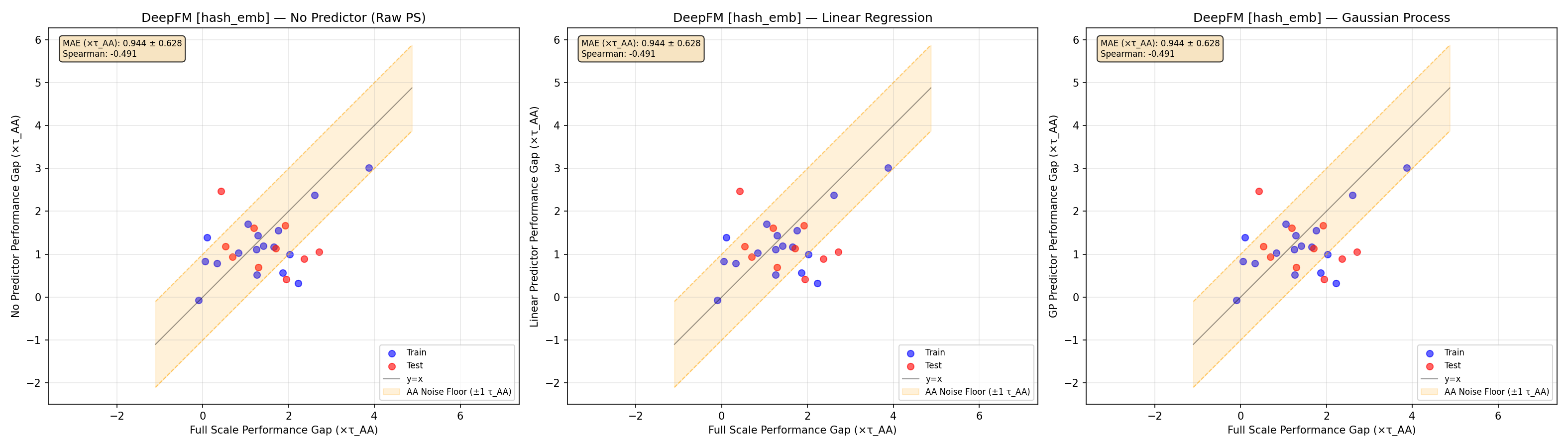}
\caption{Hash + embedding scaling}
\label{fig:deepfm-hash-emb}
\end{subfigure}
\begin{subfigure}[b]{0.65\textwidth}
\centering
\includegraphics[width=\textwidth]{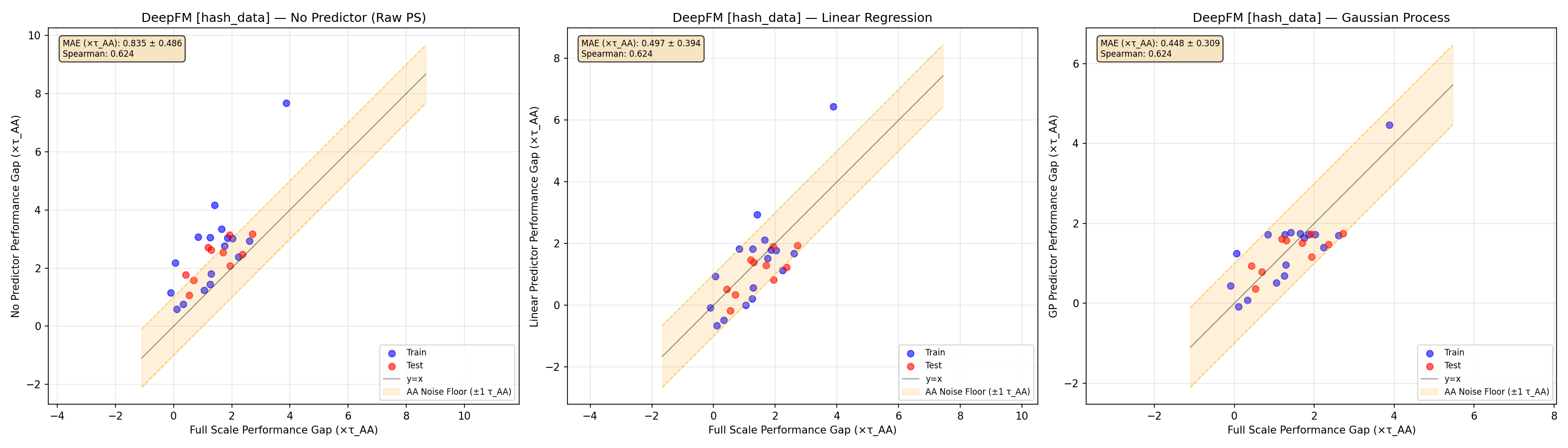}
\caption{Hash + data scaling}
\label{fig:deepfm-hash-data}
\end{subfigure}
\begin{subfigure}[b]{0.65\textwidth}
\centering
\includegraphics[width=\textwidth]{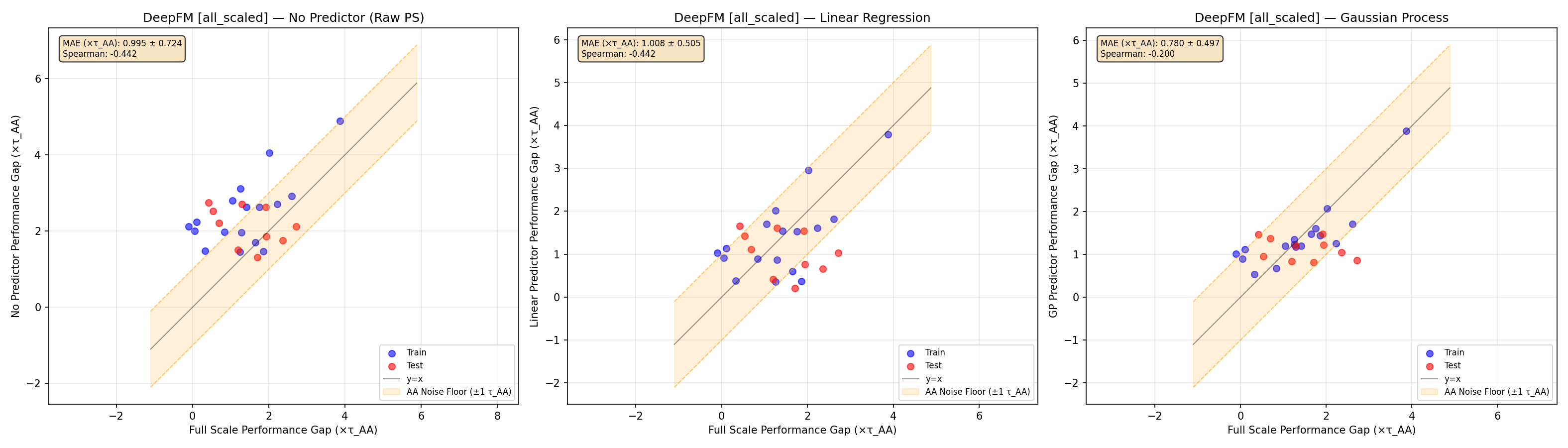}
\caption{All dimensions scaled}
\label{fig:deepfm-all-scaled}
\end{subfigure}
\caption{Predicted vs.\ actual accuracy gap for six scaling configurations on the public DeepFM benchmark. Each point represents one hypothesis; the dashed line indicates perfect prediction. Tighter clustering around the diagonal indicates higher proxy fidelity.}
\label{fig:deepfm-scatter-all}
\end{figure*}

\begin{figure*}[t]
\centering
\begin{subfigure}[b]{0.65\textwidth}
\centering
\includegraphics[width=\textwidth]{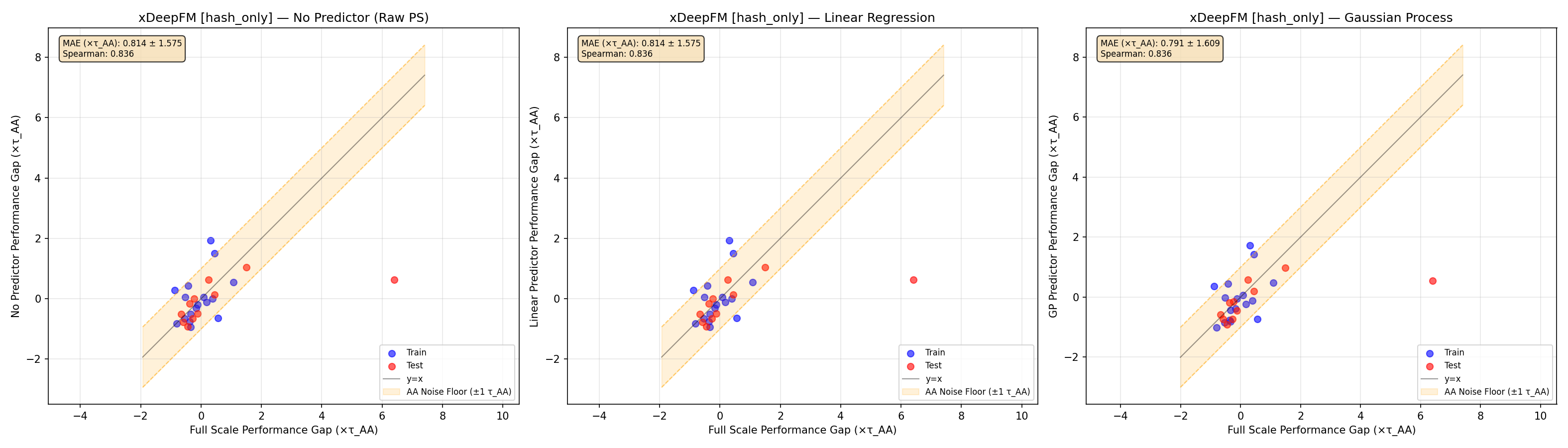}
\caption{Hash-only scaling}
\label{fig:xdeepfm-hash-only}
\end{subfigure}
\begin{subfigure}[b]{0.65\textwidth}
\centering
\includegraphics[width=\textwidth]{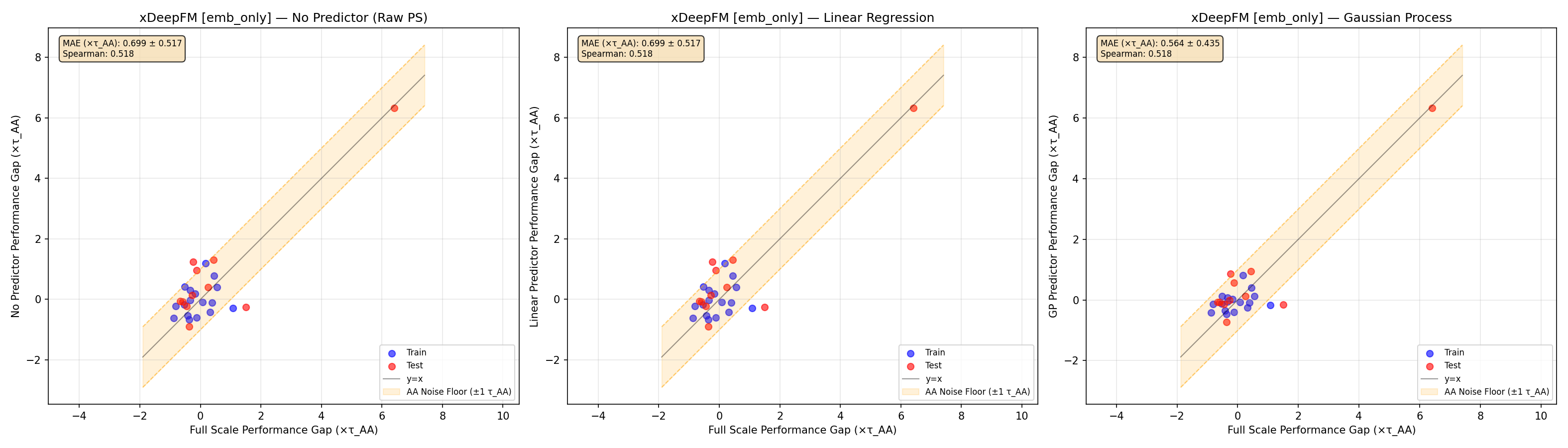}
\caption{Embedding-only scaling}
\label{fig:xdeepfm-emb-only}
\end{subfigure}
\begin{subfigure}[b]{0.65\textwidth}
\centering
\includegraphics[width=\textwidth]{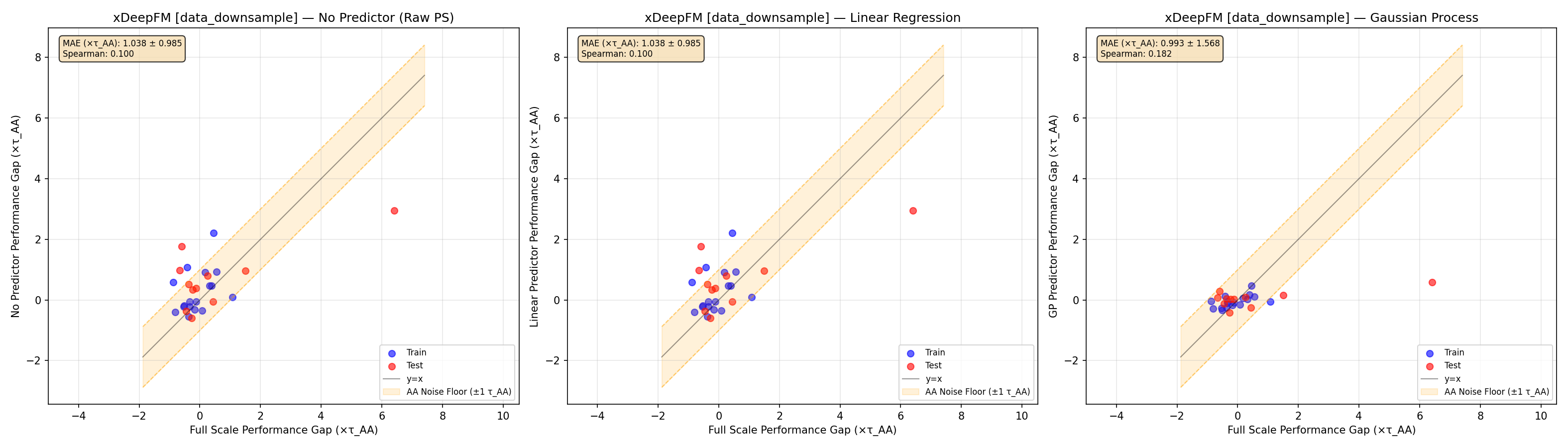}
\caption{Data downsampling}
\label{fig:xdeepfm-data-ds}
\end{subfigure}
\begin{subfigure}[b]{0.65\textwidth}
\centering
\includegraphics[width=\textwidth]{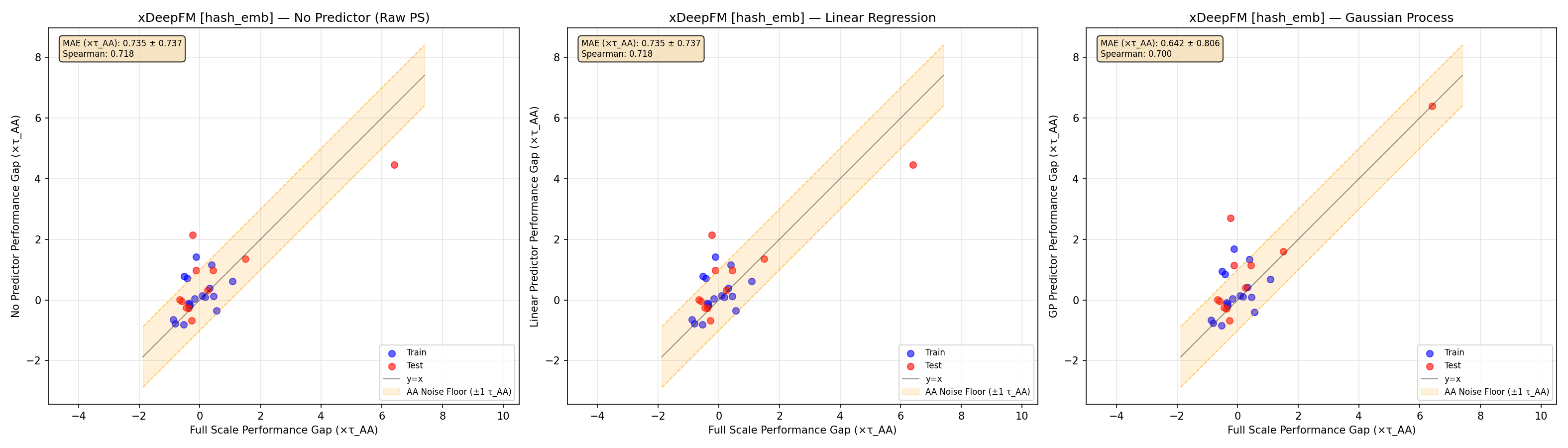}
\caption{Hash + embedding scaling}
\label{fig:xdeepfm-hash-emb}
\end{subfigure}
\begin{subfigure}[b]{0.65\textwidth}
\centering
\includegraphics[width=\textwidth]{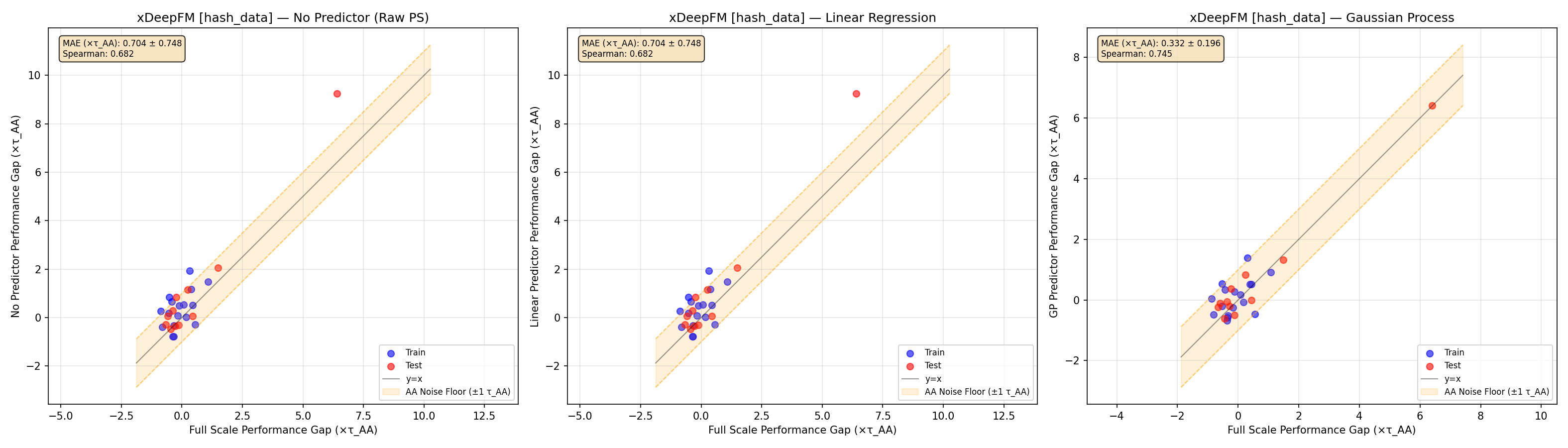}
\caption{Hash + data scaling}
\label{fig:xdeepfm-hash-data}
\end{subfigure}
\begin{subfigure}[b]{0.65\textwidth}
\centering
\includegraphics[width=\textwidth]{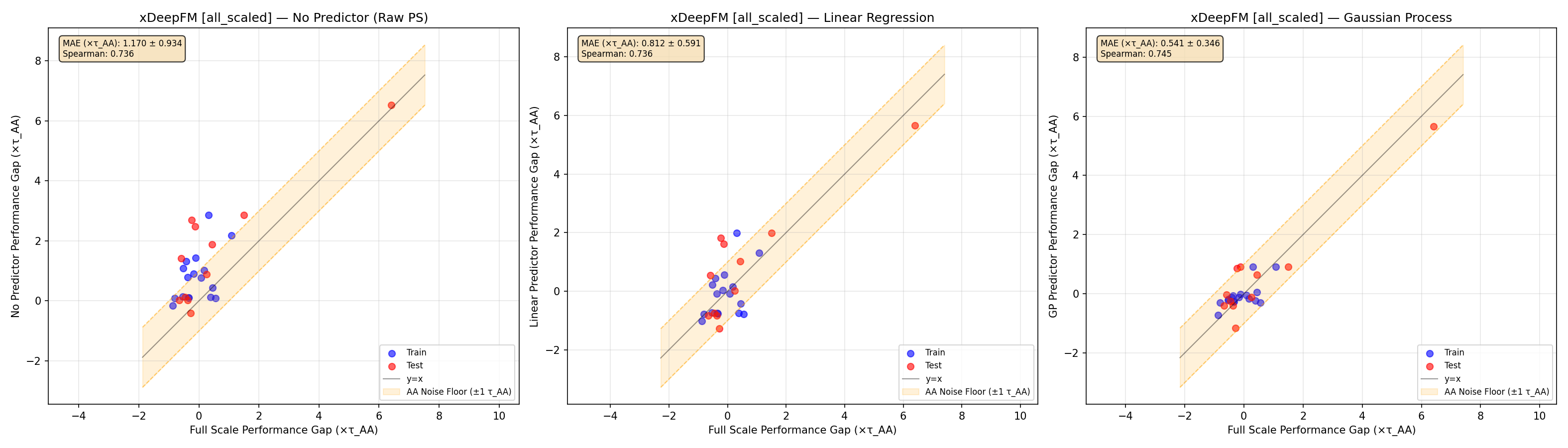}
\caption{All dimensions scaled}
\label{fig:xdeepfm-all-scaled}
\end{subfigure}
\caption{Predicted vs.\ actual accuracy gap for six scaling configurations on the public xDeepFM benchmark. Each point represents one hypothesis; the dashed line indicates perfect prediction. Tighter clustering around the diagonal indicates higher proxy fidelity.}
\label{fig:xdeepfm-scatter-all}
\end{figure*}

\begin{figure*}[t]
\centering
\begin{subfigure}[b]{0.65\textwidth}
\centering
\includegraphics[width=\textwidth]{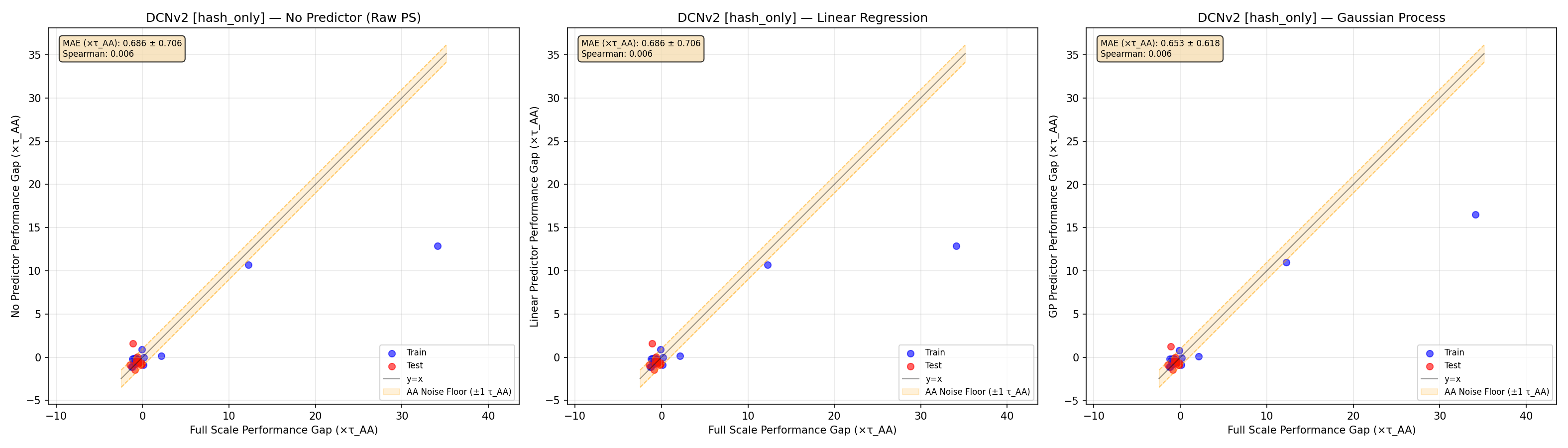}
\caption{Hash-only scaling}
\label{fig:dcnv2-hash-only}
\end{subfigure}
\begin{subfigure}[b]{0.65\textwidth}
\centering
\includegraphics[width=\textwidth]{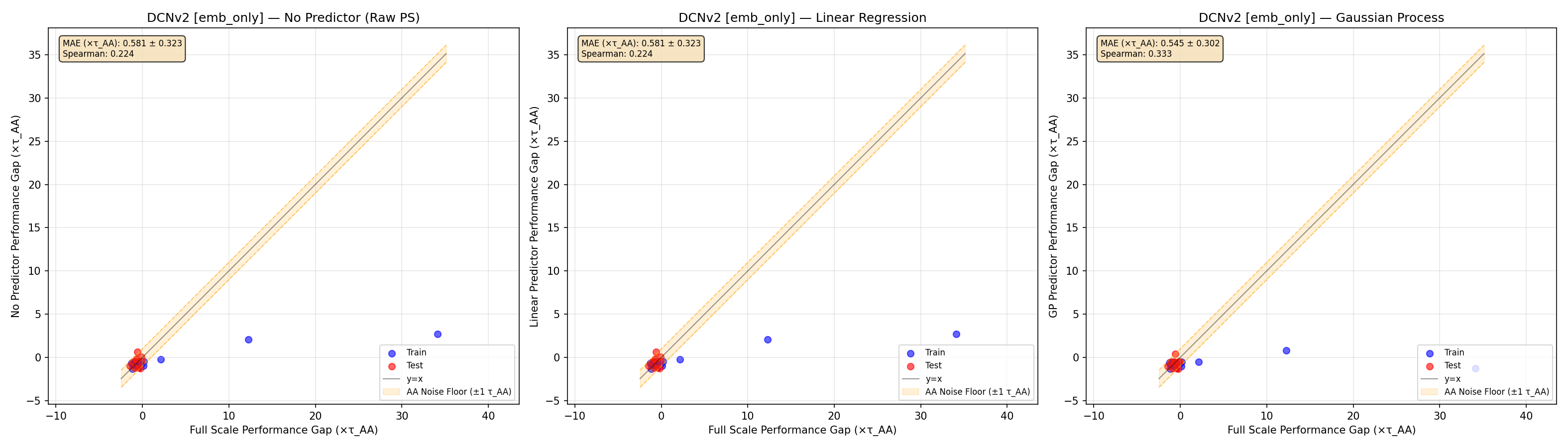}
\caption{Embedding-only scaling}
\label{fig:dcnv2-emb-only}
\end{subfigure}
\begin{subfigure}[b]{0.65\textwidth}
\centering
\includegraphics[width=\textwidth]{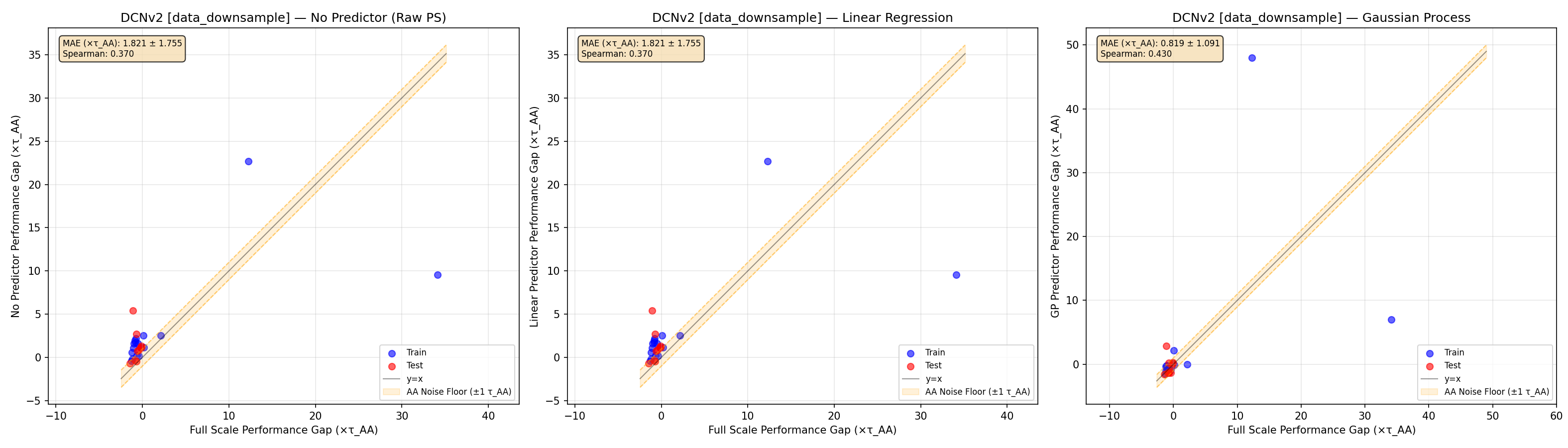}
\caption{Data downsampling}
\label{fig:dcnv2-data-ds}
\end{subfigure}
\begin{subfigure}[b]{0.65\textwidth}
\centering
\includegraphics[width=\textwidth]{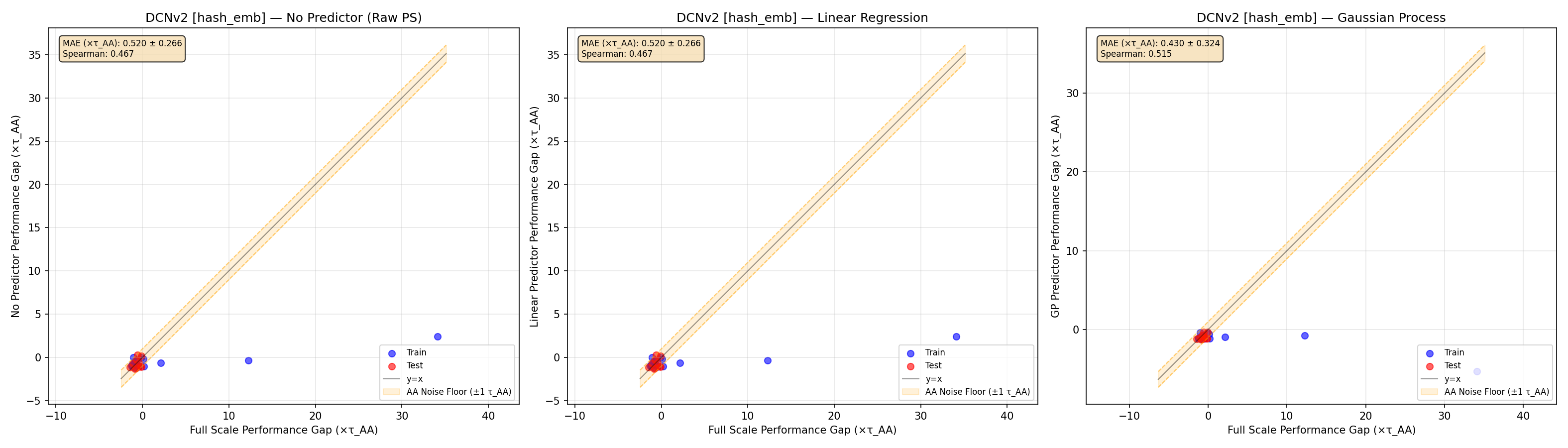}
\caption{Hash + embedding scaling}
\label{fig:dcnv2-hash-emb}
\end{subfigure}
\begin{subfigure}[b]{0.65\textwidth}
\centering
\includegraphics[width=\textwidth]{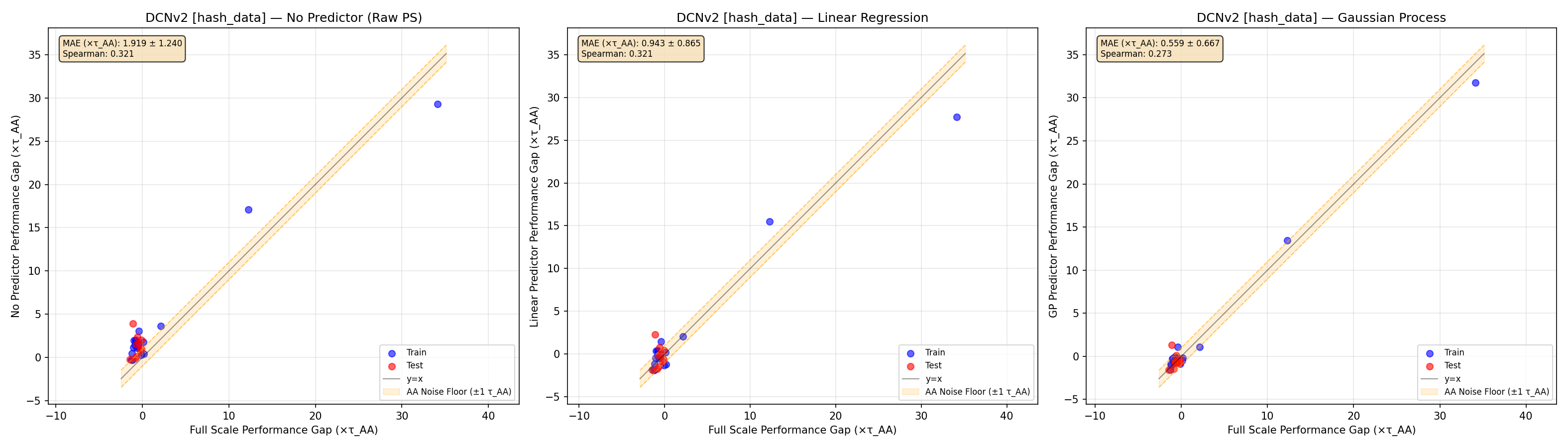}
\caption{Hash + data scaling}
\label{fig:dcnv2-hash-data}
\end{subfigure}
\begin{subfigure}[b]{0.65\textwidth}
\centering
\includegraphics[width=\textwidth]{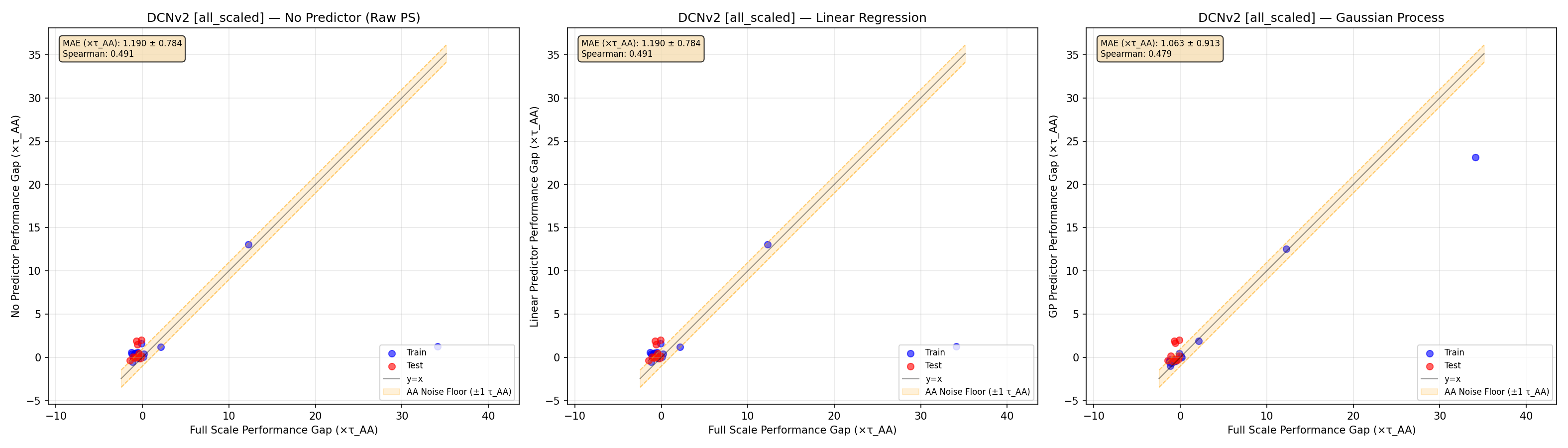}
\caption{All dimensions scaled}
\label{fig:dcnv2-all-scaled}
\end{subfigure}
\caption{Predicted vs.\ actual accuracy gap for six scaling configurations on the public DCN-v2 benchmark. Each point represents one hypothesis; the dashed line indicates perfect prediction. Tighter clustering around the diagonal indicates higher proxy fidelity.}
\label{fig:dcnv2-scatter-all}
\end{figure*}

\paragraph{Consistent MAE recovery across architectures.} The GP predictor consistently reduces MAE relative to the raw proxy, with the largest improvements occurring precisely when the raw proxy is most degraded---the regime where PS adds the most practical value. Across all four models, when the raw proxy exceeds $1\,\tau_{\text{AA}}$, the GP reduces MAE by $37$--$80\%$:

\begin{itemize}
\item \textbf{DLRM} hash\_data: $4.567 \to 0.893\,\tau_{\text{AA}}$ ($-80\%$; \cref{fig:dlrm-hash-data})
\item \textbf{DCN-v2} hash\_data: $1.919 \to 0.559\,\tau_{\text{AA}}$ ($-71\%$; \cref{fig:dcnv2-hash-data})
\item \textbf{DCN-v2} data\_ds: $1.821 \to 0.819\,\tau_{\text{AA}}$ ($-55\%$; \cref{fig:dcnv2-data-ds})
\item \textbf{xDeepFM} all\_scaled: $1.170 \to 0.541\,\tau_{\text{AA}}$ ($-54\%$; \cref{fig:xdeepfm-all-scaled})
\item \textbf{xDeepFM} hash\_data: $0.704 \to 0.332\,\tau_{\text{AA}}$ ($-53\%$; \cref{fig:xdeepfm-hash-data})
\item \textbf{DeepFM} hash\_data: $0.835 \to 0.448\,\tau_{\text{AA}}$ ($-46\%$; \cref{fig:deepfm-hash-data})
\item \textbf{DLRM} data\_ds: $0.781 \to 0.491\,\tau_{\text{AA}}$ ($-37\%$; \cref{fig:dlrm-data-ds})
\end{itemize}

\noindent This pattern---the GP providing the largest improvements where the raw proxy is worst---directly validates the finding from \cref{sec:results} that curve fitting adds the most value when the raw signal is noisy, and least when the scaling function alone already satisfies \textbf{(A2)}--\textbf{(A3)}. For configurations where the raw proxy is already accurate (e.g., DeepFM hash\_only: $0.381\,\tau_{\text{AA}}$; DLRM hash\_only: $0.731\,\tau_{\text{AA}}$), the GP provides negligible or modest additional benefit, as expected.

The GP also improves rank preservation ($\rho$) in several cases where it matters most: DLRM data\_ds ($+0.250$), DeepFM all\_scaled ($+0.242$), DLRM hash\_data ($+0.167$), and xDeepFM hash\_data ($+0.063$).

\paragraph{Linear predictor limitations.} The linear predictor adds value on DLRM (improving MAE on 5/6 configurations) but is identical to the raw proxy on 5/6 configurations for both xDeepFM and DCN-v2. This suggests that the gap transfer function $f^*$ (\cref{sec:framework}) is approximately affine for DLRM's simple dot-product interactions but nonlinear for models with explicit feature-crossing layers (CIN in xDeepFM, cross-network in DCN-v2). Rather than weakening the case for PS, this validates the GP as the default predictor choice: the GP subsumes the linear predictor and additionally captures nonlinear gap relationships through its empirical prior (\cref{sec:extrapolation}).

\subsubsection{Scaling Function Effectiveness and Architecture Dependence}
\label{app:public-scaling-effectiveness}

\paragraph{Best configurations vary by architecture.} No single scaling function dominates across all four models, directly validating the budget-optimal proxy selection procedure (\cref{alg:joint-selection}, \cref{sec:proxy-selection}):

\begin{center}
\begin{tabular}{@{}llcc@{}}
\toprule
\textbf{Model} & \textbf{Best config (GP)} & \textbf{GP MAE ($\times\tau_{\text{AA}}$)} & \textbf{GP $\rho$} \\
\midrule
DLRM & data\_ds & $0.491$ & $0.783$ \\
DeepFM & hash\_only & $0.285$ & $0.939$ \\
xDeepFM & hash\_data & $0.332$ & $0.745$ \\
DCN-v2 & hash\_emb & $0.430$ & $0.515$ \\
\bottomrule
\end{tabular}
\end{center}

\noindent This architecture dependence is a central finding: blindly applying a fixed scaling configuration across models would select a suboptimal proxy for at least three of four architectures. The PS framework addresses this through per-model selection, and the public benchmark provides empirical evidence that this selection is necessary.

\paragraph{Strong proxy fidelity is achievable.} Under the best GP configuration, all four models achieve MAE well below the noise floor: $0.285$--$0.491\,\tau_{\text{AA}}$ for DLRM, DeepFM, and xDeepFM, and $0.430\,\tau_{\text{AA}}$ for DCN-v2. The top-performing configurations also achieve meaningful rank preservation ($\rho = 0.515$--$0.939$). These results demonstrate that PS generalizes beyond our full scale models to architecturally diverse public models at $50\%$ compute reduction per scaling axis.

\paragraph{Data downsampling: robust on DLRM, variable elsewhere.} In the evaluation (\cref{sec:results}), data downsampling (DD) was the most reliable scaling function for order preservation. On DLRM, this finding replicates strongly (GP: $\rho = 0.783$, MAE $= 0.491\,\tau_{\text{AA}}$). However, DD achieves only moderate rank preservation on DeepFM ($\rho = 0.430$) and DCN-v2 ($\rho = 0.430$), and near-random rankings on xDeepFM ($\rho = 0.182$). We attribute this discrepancy to dataset scale: full scale models train on $\mathcal{O}(10^{10})$--$\mathcal{O}(10^{11})$ examples, where $50\%$ downsampling still retains billions of samples with minimal distributional shift. On the smaller Criteo dataset, $50\%$ downsampling may remove sufficient signal to degrade proxy fidelity for architectures with complex feature interactions (CIN layers in xDeepFM, cross-network layers in DCN-v2). This highlights that DD's robustness is scale-dependent, and practitioners should verify DD effectiveness on their specific data volume.

\subsubsection{Regularity Condition Violations and Failure Modes}
\label{app:public-failures}

\paragraph{Embedding scaling violates (A3) across architectures.} Embedding-only scaling (ES) produces poor rank preservation on all four models: DLRM ($\rho = -0.050$), DeepFM ($\rho = 0.115$), xDeepFM ($\rho = 0.518$), and DCN-v2 ($\rho = 0.224$ raw, $0.333$ GP). Even the GP cannot repair the rank ordering when embedding scaling is the sole scaling axis---the \textbf{(A3)} violation is structural rather than correctable by curve fitting. This contrasts with the results where ES performs well on TS-1, and likely reflects that full scale models' deeper interaction networks are more robust to embedding dimension changes than these public architectures' simpler structures. The failure is detectable during the proxy selection phase (\cref{alg:joint-selection}), where configs with poor benchmark-corpus $\rho$ are automatically excluded.

\paragraph{Multi-axis compositions can degrade fidelity.} Combining scaling dimensions sometimes introduces interaction effects that violate \textbf{(A3)} more severely than either axis alone:

\begin{itemize}
\item \textbf{DeepFM} hash\_emb ($\rho = -0.491$) combines hash-only ($\rho = 0.939$) and emb-only ($\rho = 0.115$)---the composition is dramatically worse than either component. Simultaneously reducing embedding table rows and embedding vector widths creates a compounding distortion in the factorization machine layer.
\item \textbf{DeepFM} all\_scaled ($\rho = -0.442$ raw) is similarly degraded, though the GP partially recovers ($\rho = -0.200$).
\end{itemize}

\noindent These interaction effects validate the PS design principle that scaling functions should not be composed blindly: the compute budget-optimal selection procedure (\cref{sec:proxy-selection}) evaluates compositions on a benchmark corpus and rejects those that degrade proxy fidelity.

\paragraph{DCN-v2 presents the hardest case.} DCN-v2 is the most challenging architecture for PS in this benchmark. Hash-only scaling produces essentially random rankings ($\rho = 0.006$) despite achieving reasonable MAE ($0.653\,\tau_{\text{AA}}$ under GP), indicating that MAE and rank preservation can diverge---a systematic bias can produce low MAE while scrambling relative ordering. The best GP $\rho$ across all configs is $0.515$ (hash\_emb), substantially below the other models. On all\_scaled, the GP MAE ($1.063\,\tau_{\text{AA}}$) exceeds the noise floor, and the GP slightly degrades $\rho$ on both hash\_data ($0.321 \to 0.273$) and all\_scaled ($0.491 \to 0.479$).

We attribute DCN-v2's difficulty to its cross-network layers, which create tightly coupled multiplicative interactions across all feature dimensions. The uniform $\alpha = 0.5$ scaling used in this benchmark may be too aggressive for DCN-v2's cross-network rank---analogous to the finding in \cref{app:es-vs-was} that tightly coupled full scale models (FM-1) require WAS rather than ES. This motivates architecture-aware scaling ratio selection as future work: rather than applying a uniform factor, the scaling ratio could be adapted based on architectural coupling strength.

\subsubsection{Cross-Architecture Synthesis}
\label{app:public-synthesis}

The public benchmark validates the paper's three core claims while revealing important boundary conditions:

\paragraph{Validated claims.}
\begin{enumerate}
\item \textbf{GP consistently improves over raw proxy.} The GP reduces MAE on all 24 model$\times$config pairs, with improvements ranging from $3\%$ (xDeepFM hash\_only) to $80\%$ (DLRM hash\_data).
\item \textbf{Per-model scaling function selection is essential.} The best configuration differs across all four architectures (data\_ds, hash\_only, hash\_data, hash\_emb), and applying any single configuration universally would select a suboptimal proxy for at least three models.
\item \textbf{Regularity violations are detectable.} Configurations that violate \textbf{(A3)} (negative $\rho$) are identifiable on the benchmark corpus and would be excluded by \cref{alg:joint-selection}, preventing their use on new hypotheses.
\end{enumerate}

\paragraph{Boundary conditions and limitations.}
\begin{enumerate}
\item \textbf{Architecture sensitivity.} Models with tightly coupled feature interactions (DCN-v2's cross-network, xDeepFM's CIN) are harder to scale than models with simpler interaction structures (DLRM). The GP cannot fully compensate when the scaling function itself destroys the signal.
\item \textbf{Embedding scaling is unreliable.} ES produces poor rank preservation across all four public architectures, in contrast to its effectiveness on full scale models. This suggests ES is best suited for models with deep, redundant embedding representations rather than the relatively compact embeddings in these public benchmarks.
\item \textbf{Data downsampling is scale-dependent.} DD's robustness on data ($\mathcal{O}(10^{10})$ examples) does not fully transfer to the smaller Criteo dataset, where $50\%$ downsampling may remove too much signal for architectures with complex feature interactions.
\item \textbf{Linear predictor has narrow applicability.} The linear predictor is effective only when the gap transfer function is approximately affine (DLRM), and is superseded by the GP on all architectures---supporting the GP as the default predictor choice.
\end{enumerate}

\noindent Taken together, these results demonstrate that PS generalizes across four architecturally diverse public models, with the GP predictor and per-model scaling function selection jointly achieving GP MAE of $0.285$--$0.541\,\tau_{\text{AA}}$ on the best configuration for three of four models. The fourth (DCN-v2) achieves $0.430\,\tau_{\text{AA}}$ MAE but with limited rank preservation ($\rho = 0.515$), identifying architecture-aware scaling ratios as a key direction for future work.

\begin{table}[t]
\centering
\resizebox{\textwidth}{!}{%
\begin{tabular}{@{}llccrrc@{}}
\toprule
\textbf{Config} & \textbf{Predictor} & \textbf{MAE ($\times\tau_{\text{AA}}$)} & \textbf{SE ($\times\tau_{\text{AA}}$)} & \textbf{$\Delta$MAE} & \textbf{Spearman $\rho$} & \textbf{$\Delta\rho$} \\
\midrule
\multirow{3}{*}{hash\_only}
  & Raw proxy & $0.731$ & $\pm0.617$ & --- & $0.417$ & --- \\
  & Linear & $0.731$ & $\pm0.617$ & $0\%$ & $0.417$ & $0.000$ \\
  & GP (EGP) & $0.728$ & $\pm0.479$ & $0\%$ & $0.417$ & $0.000$ \\
\midrule
\multirow{3}{*}{emb\_only}
  & Raw proxy & $1.174$ & $\pm0.494$ & --- & $-0.050$ & --- \\
  & Linear & $0.991$ & $\pm0.875$ & $-16\%$ & $-0.050$ & $0.000$ \\
  & GP (EGP) & $0.947$ & $\pm0.771$ & $-19\%$ & $-0.050$ & $0.000$ \\
\midrule
\multirow{3}{*}{data\_ds}
  & Raw proxy & $0.781$ & $\pm1.056$ & --- & $0.533$ & --- \\
  & Linear & $0.523$ & $\pm0.397$ & $-33\%$ & $0.533$ & $0.000$ \\
  & GP (EGP) & $\bm{0.491}$ & $\pm0.407$ & $-37\%$ & $\bm{0.783}$ & $+0.250$ \\
\midrule
\multirow{3}{*}{hash\_emb}
  & Raw proxy & $1.349$ & $\pm1.474$ & --- & $-0.050$ & --- \\
  & Linear & $1.219$ & $\pm1.217$ & $-10\%$ & $-0.050$ & $0.000$ \\
  & GP (EGP) & $1.219$ & $\pm1.217$ & $-10\%$ & $-0.050$ & $0.000$ \\
\midrule
\multirow{3}{*}{hash\_data}
  & Raw proxy & $4.567$ & $\pm0.901$ & --- & $0.483$ & --- \\
  & Linear & $0.921$ & $\pm1.051$ & $-80\%$ & $0.483$ & $0.000$ \\
  & GP (EGP) & $0.893$ & $\pm1.055$ & $-80\%$ & $0.650$ & $+0.167$ \\
\midrule
\multirow{3}{*}{all\_scaled}
  & Raw proxy & $0.970$ & $\pm0.545$ & --- & $0.750$ & --- \\
  & Linear & $0.852$ & $\pm0.805$ & $-12\%$ & $0.750$ & $0.000$ \\
  & GP (EGP) & $0.852$ & $\pm0.805$ & $-12\%$ & $0.750$ & $0.000$ \\
\bottomrule
\end{tabular}
}%
\caption{Public DLRM prediction accuracy across six scaling configurations. MAE and SE reported as multiples of $\tau_{\text{AA}}$. $\Delta$MAE: relative \% change vs.\ raw proxy; $\Delta\rho$: absolute difference vs.\ raw proxy.}
\label{tab:public-dlrm-all}
\end{table}

\begin{table}[t]
\centering
\resizebox{\textwidth}{!}{%
\begin{tabular}{@{}llccrrc@{}}
\toprule
\textbf{Config} & \textbf{Predictor} & \textbf{MAE ($\times\tau_{\text{AA}}$)} & \textbf{SE ($\times\tau_{\text{AA}}$)} & \textbf{$\Delta$MAE} & \textbf{Spearman $\rho$} & \textbf{$\Delta\rho$} \\
\midrule
\multirow{3}{*}{hash\_only}
  & Raw proxy & $0.381$ & $\pm0.303$ & --- & $0.939$ & --- \\
  & Linear & $0.381$ & $\pm0.303$ & $0\%$ & $0.939$ & $0.000$ \\
  & GP (EGP) & $\bm{0.285}$ & $\pm0.304$ & $-25\%$ & $\bm{0.939}$ & $0.000$ \\
\midrule
\multirow{3}{*}{emb\_only}
  & Raw proxy & $0.848$ & $\pm0.487$ & --- & $0.115$ & --- \\
  & Linear & $0.848$ & $\pm0.487$ & $0\%$ & $0.115$ & $0.000$ \\
  & GP (EGP) & $0.795$ & $\pm0.460$ & $-6\%$ & $0.115$ & $0.000$ \\
\midrule
\multirow{3}{*}{data\_ds}
  & Raw proxy & $0.883$ & $\pm0.569$ & --- & $0.430$ & --- \\
  & Linear & $0.754$ & $\pm0.403$ & $-15\%$ & $0.430$ & $0.000$ \\
  & GP (EGP) & $0.641$ & $\pm0.327$ & $-27\%$ & $0.430$ & $0.000$ \\
\midrule
\multirow{3}{*}{hash\_emb}
  & Raw proxy & $0.944$ & $\pm0.628$ & --- & $-0.491$ & --- \\
  & Linear & $0.944$ & $\pm0.628$ & $0\%$ & $-0.491$ & $0.000$ \\
  & GP (EGP) & $0.944$ & $\pm0.628$ & $0\%$ & $-0.491$ & $0.000$ \\
\midrule
\multirow{3}{*}{hash\_data}
  & Raw proxy & $0.835$ & $\pm0.486$ & --- & $0.624$ & --- \\
  & Linear & $0.497$ & $\pm0.394$ & $-40\%$ & $0.624$ & $0.000$ \\
  & GP (EGP) & $0.448$ & $\pm0.309$ & $-46\%$ & $0.624$ & $0.000$ \\
\midrule
\multirow{3}{*}{all\_scaled}
  & Raw proxy & $0.995$ & $\pm0.724$ & --- & $-0.442$ & --- \\
  & Linear & $1.008$ & $\pm0.505$ & $+1\%$ & $-0.442$ & $0.000$ \\
  & GP (EGP) & $0.780$ & $\pm0.497$ & $-22\%$ & $-0.200$ & $+0.242$ \\
\bottomrule
\end{tabular}
}%
\caption{Public DeepFM prediction accuracy across six scaling configurations. MAE and SE reported as multiples of $\tau_{\text{AA}}$. $\Delta$MAE: relative \% change vs.\ raw proxy; $\Delta\rho$: absolute difference vs.\ raw proxy.}
\label{tab:public-deepfm-all}
\end{table}

\begin{table}[t]
\centering
\resizebox{\textwidth}{!}{%
\begin{tabular}{@{}llccrrc@{}}
\toprule
\textbf{Config} & \textbf{Predictor} & \textbf{MAE ($\times\tau_{\text{AA}}$)} & \textbf{SE ($\times\tau_{\text{AA}}$)} & \textbf{$\Delta$MAE} & \textbf{Spearman $\rho$} & \textbf{$\Delta\rho$} \\
\midrule
\multirow{3}{*}{hash\_only}
  & Raw proxy & $0.814$ & $\pm1.575$ & --- & $0.836$ & --- \\
  & Linear & $0.814$ & $\pm1.575$ & $0\%$ & $0.836$ & $0.000$ \\
  & GP (EGP) & $0.791$ & $\pm1.609$ & $-3\%$ & $0.836$ & $0.000$ \\
\midrule
\multirow{3}{*}{emb\_only}
  & Raw proxy & $0.699$ & $\pm0.517$ & --- & $0.518$ & --- \\
  & Linear & $0.699$ & $\pm0.517$ & $0\%$ & $0.518$ & $0.000$ \\
  & GP (EGP) & $0.564$ & $\pm0.435$ & $-19\%$ & $0.518$ & $0.000$ \\
\midrule
\multirow{3}{*}{data\_ds}
  & Raw proxy & $1.038$ & $\pm0.985$ & --- & $0.100$ & --- \\
  & Linear & $1.038$ & $\pm0.985$ & $0\%$ & $0.100$ & $0.000$ \\
  & GP (EGP) & $0.993$ & $\pm1.568$ & $-4\%$ & $0.182$ & $+0.082$ \\
\midrule
\multirow{3}{*}{hash\_emb}
  & Raw proxy & $0.735$ & $\pm0.737$ & --- & $0.718$ & --- \\
  & Linear & $0.735$ & $\pm0.737$ & $0\%$ & $0.718$ & $0.000$ \\
  & GP (EGP) & $0.642$ & $\pm0.806$ & $-13\%$ & $0.700$ & $-0.018$ \\
\midrule
\multirow{3}{*}{hash\_data}
  & Raw proxy & $0.704$ & $\pm0.748$ & --- & $0.682$ & --- \\
  & Linear & $0.704$ & $\pm0.748$ & $0\%$ & $0.682$ & $0.000$ \\
  & GP (EGP) & $\bm{0.332}$ & $\pm0.196$ & $-53\%$ & $\bm{0.745}$ & $+0.063$ \\
\midrule
\multirow{3}{*}{all\_scaled}
  & Raw proxy & $1.170$ & $\pm0.934$ & --- & $0.736$ & --- \\
  & Linear & $0.812$ & $\pm0.591$ & $-31\%$ & $0.736$ & $0.000$ \\
  & GP (EGP) & $0.541$ & $\pm0.346$ & $-54\%$ & $0.745$ & $+0.009$ \\
\bottomrule
\end{tabular}
}%
\caption{Public xDeepFM prediction accuracy across six scaling configurations. MAE and SE reported as multiples of $\tau_{\text{AA}}$. $\Delta$MAE: relative \% change vs.\ raw proxy; $\Delta\rho$: absolute difference vs.\ raw proxy.}
\label{tab:public-xdeepfm-all}
\end{table}

\begin{table}[t]
\centering
\resizebox{\textwidth}{!}{%
\begin{tabular}{@{}llccrrc@{}}
\toprule
\textbf{Config} & \textbf{Predictor} & \textbf{MAE ($\times\tau_{\text{AA}}$)} & \textbf{SE ($\times\tau_{\text{AA}}$)} & \textbf{$\Delta$MAE} & \textbf{Spearman $\rho$} & \textbf{$\Delta\rho$} \\
\midrule
\multirow{3}{*}{hash\_only}
  & Raw proxy & $0.686$ & $\pm0.706$ & --- & $0.006$ & --- \\
  & Linear & $0.686$ & $\pm0.706$ & $0\%$ & $0.006$ & $0.000$ \\
  & GP (EGP) & $0.653$ & $\pm0.618$ & $-5\%$ & $0.006$ & $0.000$ \\
\midrule
\multirow{3}{*}{emb\_only}
  & Raw proxy & $0.581$ & $\pm0.323$ & --- & $0.224$ & --- \\
  & Linear & $0.581$ & $\pm0.323$ & $0\%$ & $0.224$ & $0.000$ \\
  & GP (EGP) & $0.545$ & $\pm0.302$ & $-6\%$ & $0.333$ & $+0.109$ \\
\midrule
\multirow{3}{*}{data\_ds}
  & Raw proxy & $1.821$ & $\pm1.755$ & --- & $0.370$ & --- \\
  & Linear & $1.821$ & $\pm1.755$ & $0\%$ & $0.370$ & $0.000$ \\
  & GP (EGP) & $0.819$ & $\pm1.091$ & $-55\%$ & $0.430$ & $+0.060$ \\
\midrule
\multirow{3}{*}{hash\_emb}
  & Raw proxy & $0.520$ & $\pm0.266$ & --- & $0.467$ & --- \\
  & Linear & $0.520$ & $\pm0.266$ & $0\%$ & $0.467$ & $0.000$ \\
  & GP (EGP) & $\bm{0.430}$ & $\pm0.324$ & $-17\%$ & $\bm{0.515}$ & $+0.048$ \\
\midrule
\multirow{3}{*}{hash\_data}
  & Raw proxy & $1.919$ & $\pm1.240$ & --- & $0.321$ & --- \\
  & Linear & $0.943$ & $\pm0.865$ & $-51\%$ & $0.321$ & $0.000$ \\
  & GP (EGP) & $0.559$ & $\pm0.667$ & $-71\%$ & $0.273$ & $-0.048$ \\
\midrule
\multirow{3}{*}{all\_scaled}
  & Raw proxy & $1.190$ & $\pm0.784$ & --- & $0.491$ & --- \\
  & Linear & $1.190$ & $\pm0.784$ & $0\%$ & $0.491$ & $0.000$ \\
  & GP (EGP) & $1.063$ & $\pm0.913$ & $-11\%$ & $0.479$ & $-0.012$ \\
\bottomrule
\end{tabular}
}%
\caption{Public DCN-v2 prediction accuracy across six scaling configurations. MAE and SE reported as multiples of $\tau_{\text{AA}}$. $\Delta$MAE: relative \% change vs.\ raw proxy; $\Delta\rho$: absolute difference vs.\ raw proxy.}
\label{tab:public-dcnv2-all}
\end{table}

\end{document}